\documentclass[lettersize,journal]{IEEEtran}
\usepackage[T1]{fontenc}
\usepackage{array}
\usepackage{stfloats}
\usepackage{verbatim}
\usepackage{subfigure}
\usepackage{cite}
\usepackage{makecell}
\usepackage{amsmath,amssymb,amsfonts}
\usepackage{graphicx}
\usepackage{textcomp}
\usepackage{color,xcolor}
\usepackage{wrapfig}
\usepackage{hyperref}
\usepackage{booktabs}
\usepackage{amsthm}
\usepackage{url}
\usepackage{diagbox}
\usepackage{threeparttable}
\usepackage{algorithm}
\usepackage{algorithmic}

\theoremstyle{definition}

\DeclareRobustCommand{\revise}[1]{\textcolor{black}{#1}}

\begin{document}
\newcommand{\mytitle}{{\sf TriDeliver}: Cooperative Air-Ground Instant Delivery with UAVs, Couriers, and Crowdsourced Ground Vehicles}
\title{\huge \mytitle} 

\author{Junhui Gao,
        Yan Pan,
        Qianru Wang,
		Wenzhe Hou,
        Yiqin Deng,
        Liangliang Jiang,
        and Yuguang Fang,~\IEEEmembership{Fellow,~IEEE}.

\IEEEcompsocitemizethanks{\IEEEcompsocthanksitem Junhui Gao and Yuguang Fang are with Hong Kong JC STEM Lab of Smart City and Department of Computer Science, City University of Hong Kong, Kowloon, Hong Kong, China.\protect\ E-mail: junhui.gao@my.cityu.edu.hk, my.Fang@cityu.edu.hk.
\IEEEcompsocthanksitem Yan Pan is with State Key Laboratory of
Complex \& Critical Software Environment, and College of Computer Science and Technology, National University of Defense Technology, Changsha, China.\ E-mail: panyan@nudt.edu.cn.
\IEEEcompsocthanksitem Qianru Wang is with the School of Computer Science and Technology, Xidian University, Xi'an, China, and Hong Kong JC STEM Lab of Smart City and Department of Computer Science, City University of Hong Kong, Kowloon, Hong Kong, China.\ E-mail: wangqianru@xidian.edu.cn.
\IEEEcompsocthanksitem Wenzhe Hou is with National Key Laboratory of Big Data and Decision, National University of Defense Technology, China.\ E-mail: hwz241mic@hotmail.com
\IEEEcompsocthanksitem Yiqin Deng is with School of Data Science, Lingnan University, Hong Kong, China. \ E-mail: yiqindeng@ln.edu.hk.
\IEEEcompsocthanksitem Liangliang Jiang is with School of Accounting and Finance, Hong Kong Polytechnic University, Kowloon, Hong Kong, China. \ E-mail: liangliang.jiang@polyu.edu.hk
}}



\maketitle

\begin{abstract}
Instant delivery \revise{is an important urban logistics service characterized by strict} deadlines \revise{and rapidly growing demand}.
\revise{Couriers}, \revise{unmanned aerial vehicles} (UAVs), and crowdsourced \revise{ground vehicles (GVs) offer complementary capabilities}, \revise{but courier delivery is labor-intensive and faces capacity pressure as demand grows}, \revise{UAV delivery is constrained by} flight \revise{restrictions and endurance}, and \revise{crowdsourced-GV availability is unstable}.
\revise{Existing UAV--courier and UAV--GV cooperative schemes partly alleviate these limitations but remain constrained by costly dedicated courier fleets or unstable crowdsourced-GV supply.}
\revise{Therefore, this} paper proposes {\sf TriDeliver}, the first framework \revise{to jointly coordinate} couriers, UAVs, and crowdsourced GVs for \revise{city-scale} instant delivery.
To \revise{address} the \revise{cold-start problem arising from sparse UAV/GV assignment logs}, we \revise{develop} a \revise{transfer-learning-based method that adapts courier-derived parcel-assignment preferences} to \revise{the distinct operating constraints of} UAVs and GVs.
\revise{The resulting agent-specific preference functions guide cooperative assignment, followed by lexicographic optimization of the remaining parcels.}
\revise{Data-driven simulations using 30-day} real-world \revise{datasets show that, under the $100\%$ demand setting} and \revise{the default configuration}, \revise{{\sf TriDeliver} fulfills $97.6\%$ of orders.}
\revise{Compared with the state-of-the-art UAV--taxi cooperative baseline}, \revise{it delivers $25.9\%$ more parcels} and reduces \revise{total} delivery cost by \revise{$60.7\%$.}
\revise{The ablation study} further \revise{shows that transfer learning reduces average} delivery time \revise{by $5.7\%$}, \revise{total} delivery cost \revise{by $5.8\%$}, and \revise{average taxi delivery price by $20.9\%$ relative to the variant without transfer learning}, \revise{while maintaining essentially} the \revise{same fulfillment level}.
\end{abstract}

\begin{IEEEkeywords}
UAV, Crowdsourcing, Air-Ground Cooperative Delivery, Transfer Learning, Mobile Edge Computing.
\end{IEEEkeywords}

\section{Introduction}
\label{sec.introduction}


Instant delivery aims to deliver an item (e.g., food \cite{jiang2023faircod} and emergency medicine \cite{kobusingye2006emergency}) within a very short timeframe (e.g., \revise{an hour}), fulfilling a fundamental daily service. \revise{The demand supported by instant-delivery infrastructure is already substantial and continues to grow. In China, the transaction value of instant retail reached 650 billion Chinese Yuan (CNY) in 2023, representing an annual increase of 28.89\%, and is projected to exceed CNY 2 trillion by 2030~\cite{CAITECInstantRetail2024}. At the platform level, Meituan reported 21.893 billion on-demand delivery transactions in 2023, up 23.9\% year on year~\cite{MeituanAnnualReport2023}. These trends make scalable instant-delivery capacity an increasingly important urban-logistics problem.}

The growing demand \revise{for} instant delivery exposes a fundamental mismatch: no single delivery model can reliably meet \revise{peak-demand} requirements on its own.
\revise{Several major platforms have deployed UAV} delivery \revise{services on a limited scale:} JD \revise{has deployed delivery drones in selected areas~\cite{JDAnnualReport2021}; by the end of 2024}, Meituan \revise{had opened 53 UAV routes and completed more than 450,000 orders~\cite{MeituanUAV2025}; and Amazon Prime Air had begun commercial operations under FAA certification~\cite{FAAPart135DroneDelivery}. Nevertheless, these deployments remain constrained in geographical coverage, operating conditions}, and \revise{service capacity. Available} delivery approaches can be roughly divided into three categories: unmanned aerial delivery (i.e., UAV delivery)~\cite{7989675,dissanayaka2023review,pan2021efficient}, courier delivery~\revise{\cite{ding2021nationwide,10.1145/3372224.3419198}}, and \revise{crowdsourced} delivery~\revise{\cite{wang2022recommending,ding2021city,xie2023transfloor,fatehi2022crowdsourcing}}.
\revise{Each} delivery \revise{method has distinct strengths} and \revise{limitations}. Specifically, \revise{UAV} delivery is fast and flexible. However, this approach faces limitations in capacity and range due to no-fly zones and limited battery \revise{capacity}, despite being unaffected by traffic conditions~\cite{pan2024pioneering,wen2022joint,8974403}. \revise{Furthermore, recent UAV-assisted 6G architectures employ edge computing and task offloading to improve communication latency, energy efficiency, and network reliability~\cite{ALOTAIBI2025104213}. This edge-oriented paradigm also motivates limiting unnecessary centralized transmission of fine-grained operational data.
Despite its maturity and wide deployment~\cite{ding2021nationwide},} courier delivery \revise{remains labor-intensive} and \revise{susceptible to capacity pressure as demand grows. Official labor-market monitoring in China identified couriers and on-demand delivery workers among occupations experiencing recruitment shortages~\cite{MOHRSSShortage2022}}, \revise{while an empirical last-mile study shows that growing demand can overload couriers} and \revise{leave assigned services unfinished~\cite{PegadoBardayo2023}}. \revise{This capacity pressure motivates} the \revise{introduction of complementary} delivery \revise{resources.
Crowdsourced delivery opportunistically recruits} GVs \revise{such as taxis and other} private \revise{vehicles}~\cite{behrend2019exact,ding2021city}\revise{. Although} it can reduce labor costs\revise{, its available capacity is neither stable nor fully reliable}~\cite{10598004}.
\revise{Because each method has} inherent drawbacks, \revise{its standalone} performance is limited, especially \revise{at city scale}. \revise{Cooperative schemes that integrate two} delivery \revise{modes have therefore} been proposed, including \revise{UAV--courier} cooperation~\cite{pan2024pioneering,9151388} and \revise{UAV--GV} cooperation~\cite{10598004,fatehi2022crowdsourcing,11261388}.
However, \revise{pairwise schemes retain a capacity-provisioning trade-off: provisioning dedicated UAV and courier capacity for peak demand leaves resources underused off peak, whereas schemes relying} on \revise{crowdsourced} couriers \revise{or GVs face} unstable supply~\cite{pan2024pioneering,fatehi2022crowdsourcing}.

To tackle these issues, this paper proposes \revise{{\sf TriDeliver}, an air--ground} cooperative delivery \revise{framework that} integrates \revise{all} three \revise{agent types}.
{\sf TriDeliver} \revise{combines} dedicated UAVs and couriers \revise{with} crowdsourced GVs to \revise{reduce} delivery cost\revise{, maintain stable baseline capacity,} and \revise{absorb demand surges} during rush hours.
Specifically, relying on a single delivery mode often leads to either high costs or insufficient capacity. {\sf TriDeliver} addresses this \revise{limitation} by \revise{exploiting} the complementary strengths of \revise{the} three \revise{agent types}. Instead of maintaining a \revise{large} fleet of dedicated couriers \revise{solely} to meet peak \revise{demand,} which \revise{would be costly} during off-peak \revise{periods,} we \revise{use} a \revise{smaller courier fleet} to handle the stable base load. \revise{Crowdsourced GVs then provide scalable supplemental capacity} during \revise{demand surges} without \revise{the same} fixed \revise{investment}. Furthermore, \revise{UAVs can bypass} ground congestion and \revise{accelerate feasible} urgent \revise{deliveries}. This hybrid approach \revise{is designed to improve} operational efficiency\revise{, cost effectiveness,} and reliability across varying \revise{demand} conditions.

\begin{figure}[t!]
    \centering
    \includegraphics[width=0.40\textwidth]{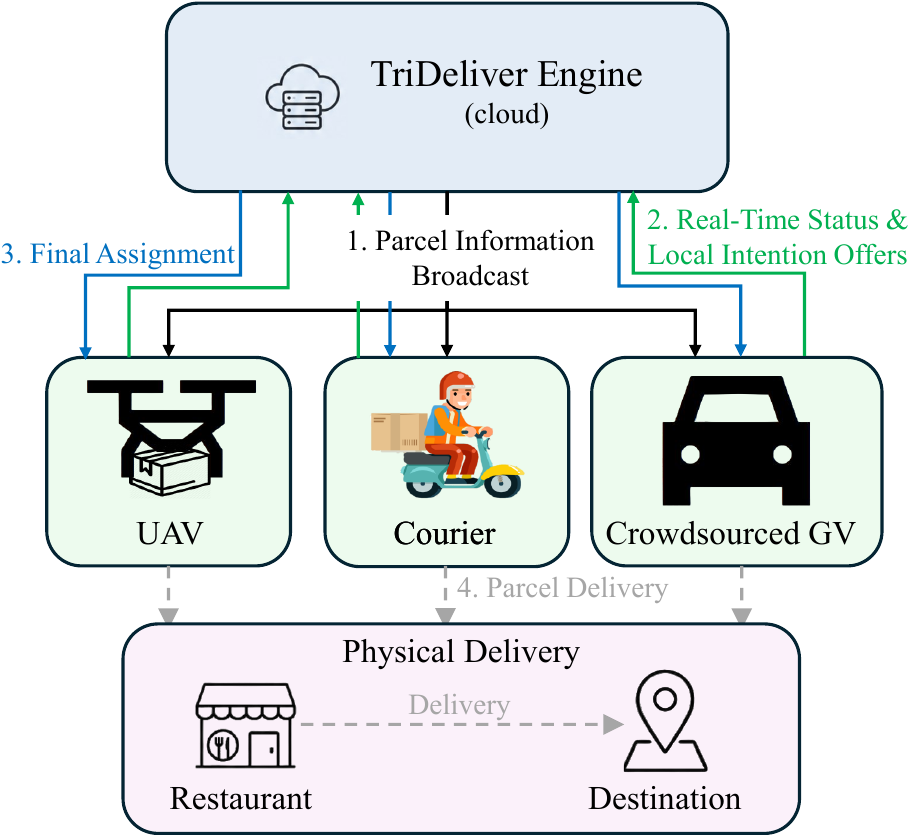}
    \vspace{-0.3cm}
    \caption{The Workflow of {\sf TriDeliver}.}
    \label{fig.architecture}
    \vspace{-0.5cm}
\end{figure}

\revise{Effective} city-scale \revise{cooperation among} UAVs, couriers, and GVs is \revise{nevertheless} challenging. \revise{City-scale} parcel assignment is commonly \revise{formulated} as an NP-hard problem~\cite{duan2023velp,10167760}, \revise{and coordinating three heterogeneous agent types introduces additional feasibility and allocation dependencies.
In particular, during the initial deployment of UAV and GV delivery services, large-scale target-domain assignment logs are typically unavailable or sparse. This creates a cold-start problem for learning-based parcel assignment, analogous to the limited-data problem faced by newly deployed smart-city services~\cite{Wang2018UrbanTransfer}. Training target-specific models from scratch would require substantial target-domain data collection and exploration, delaying effective} tripartite cooperation. \revise{Urban transfer learning provides a natural way to address this problem by transferring assignment} knowledge \revise{from a data-rich source service to data-scarce target services}. \revise{Couriers are} the most \revise{established} delivery agents, \revise{and} their \revise{parcel-selection behavior encodes domain} knowledge from both human \revise{experience} and \revise{mature} assignment algorithms~\cite{duan2023velp}. This \revise{asymmetry allows us} to \revise{learn from courier data} and \revise{fine-tune} the assignment \revise{models for} UAVs and GVs.
The \revise{transferred models support both agent-specific efficiency and system-level coordination by incorporating courier-derived} assignment \revise{knowledge into all three decision functions. Parcels} are \revise{first} assigned \revise{using these agent-specific models, and those not selected at this stage are handled} by \revise{a subsequent} optimization \revise{step}.

Fig.~\ref{fig.architecture} illustrates the workflow of {\sf TriDeliver}.
The \revise{control center broadcasts parcel information;} UAVs, couriers, and GVs \revise{then evaluate candidate deliveries locally and submit compact intention offers containing only assignment-relevant metadata}, such as \revise{the parcel}, \revise{estimated service time, marginal cost, resource margin}, and \revise{their knowledge-induced preference~\cite{10976414}.
UAV-assisted 6G architectures show that edge computing and task offloading can reduce latency and improve energy efficiency and network reliability~\cite{ALOTAIBI2025104213}.
Consistent with this edge-oriented principle}, \revise{{\sf TriDeliver} keeps route-feasibility evaluation local and avoids transmitting raw trajectories or complete routes}.
\revise{Offers are sent upon key events rather than continuously}, \revise{reducing communication load over 5G-capable low-latency links~\cite{7876982}}.
\revise{The engine selects among feasible offers, and each selected agent confirms} that \revise{its offer remains valid; GVs additionally verify route overlap, detour tolerance, and timing constraints against their private, potentially changing passenger routes.
Following} the flexible provisioning framework in~\cite{Deng_2026}\revise{, this logical dispatch} service \revise{can operate} at the network edge or \revise{in} the cloud \revise{according to} latency requirements.

The major contributions of this paper \revise{are threefold}:
\begin{itemize}
    \item {\sf TriDeliver} is the first \revise{framework to integrate} three \revise{representative} delivery methods\revise{,} UAV delivery, courier delivery, and crowdsourced delivery by GVs, \revise{and coordinate their complementary capabilities for city-scale instant} delivery.
    \item \revise{We design a transfer-learning-based parcel-assignment method that learns a courier preference function from historical courier decisions and adapts it} to UAVs and GVs \revise{through feature alignment and domain-specific fine-tuning. The three resulting preference functions support cooperative assignment}, \revise{followed by a lexicographic optimization for the remaining parcels}.
    \item \revise{Evaluations} using one-month real-world \revise{trajectory and delivery} datasets show that\revise{, under full demand and the default resource settings,} {\sf TriDeliver} \revise{fulfills $97.6\%$ of parcel orders. Compared with the state-of-the-art UAV--GV cooperative baseline, it delivers $25.9\%$ more parcels and} reduces \revise{total} delivery cost by \revise{$60.7\%$. The ablation study further shows that transfer learning reduces average delivery time by $5.7\%$, total delivery cost by $5.8\%$, and taxi delivery price by $20.9\%$ relative} to the \revise{variant without transfer learning}.
\end{itemize}

The remainder of the paper is organized as follows. 
Sec.~\ref{sec.literature} reviews the related \revise{work.} Sec.~\ref{sec.preliminary} \revise{presents the datasets}, \revise{motivation,} and framework. Sec.~\ref{sec.model} \revise{defines the delivery models}, \revise{and} Sec.~\ref{sec.TLAssignment} introduces the \revise{Transfer-Learning (TL)-based parcel-assignment strategy}. Sec.~\ref{sec.evaluation} \revise{evaluates the proposed method against the baselines, and} Sec.~\ref{sec.conclusion} concludes \revise{the} paper.

\section{Related Work}
\label{sec.literature}
In this section, we review \revise{four strands of} related \revise{work: UAV} delivery, \revise{courier delivery}, crowdsourced \revise{delivery}, and cooperative \revise{delivery}.

\revise{\textbf{UAV Delivery.}}
UAV delivery has been proposed \revise{to improve delivery efficiency}. Chen et al.~\cite{chen2022image} implemented control strategies for autonomous target \revise{pickup} by UAVs. However, handling diverse parcel shapes remains challenging. Gawel et al.~\cite{7989675} addressed this \revise{issue} with a system \revise{that picks up} parcels \revise{of varying shapes} with \revise{partially} ferrous surfaces. Path planning is another \revise{major} focus: Du et al.~\cite{du2021uav} optimized delivery paths using probabilistic \revise{geofences}, while Dorling et al.~\cite{7513397} formulated \revise{vehicle-routing} problems \revise{that} minimize delivery \revise{cost or} time. Battery \revise{capacity limits} flight range\revise{; proposed solutions include} optimized \revise{charging-station} deployment (e.g., Huang et al.~\cite{huang2020method}) and fast-charging technologies (e.g., \revise{the 500-W wireless charging system of} Cai et al.~\cite{cai2020500}). \revise{Beyond static charging infrastructure, Ameur et al.~\cite{11224006} proposed a UAV energy-management framework for 6G networks that uses UGV-mounted reconfigurable intelligent surfaces to support wireless power transfer and data relaying, thereby reducing the need for UAVs to return to the base station for energy replenishment and content reloading.}

\revise{\textbf{Courier Delivery.}}
Courier delivery \revise{is widely used in} instant parcel delivery~\revise{\cite{ding2021nationwide}}, \revise{so improving its} efficiency \revise{can} yield \revise{substantial} benefits. Sungur et al.~\cite{sungur2010model} applied stochastic programming to handle uncertainty in time-constrained daily scheduling. Chen and Hu~\cite{chen2024courier} compared dedicated \revise{and} pooling \revise{dispatch} policies, showing that \revise{the} optimal strategy depends on \revise{service-area size}, customer patience, and \revise{whether} demand \revise{can be influenced through pricing}. Parcel assignment critically \revise{affects} performance: Auad et al.~\cite{auad2024dynamic} balanced courier costs and service quality \revise{through deep Q-learning}, while Bozanta et al.~\cite{BOZANTA2022107871} evaluated three assignment strategies. For highly time-sensitive emergency delivery, Xia et al.~\cite{10.1145/3580305.3599766} \revise{studied emergency courier allocation at a scale of} over 10,000 couriers\revise{,} 170 zones\revise{, and} 100 million tasks. Zhu et al.~\cite{10184563} leveraged crowdsourced couriers alongside dedicated \revise{couriers} during peak sales. Lyu et al.~\cite{10184811} addressed long-distance delays using relay and delivery couriers across zones.

\textbf{Crowdsourced Delivery.}
Crowdsourced delivery \revise{uses} idle \revise{transport} resources\revise{, such as public-transport} passengers or taxis\revise{, to carry} shipments. Ding et al.~\cite{ding2021city} proposed time-constrained multi-hop delivery \revise{through} public transport. Liu et al.~\cite{liu2018foodnet} investigated food delivery \revise{by} crowdsourced taxis \revise{using passenger-occupancy} data. \revise{Fairness-aware} methods \revise{have} also \revise{been used to improve task-allocation ratios}~\cite{basik2018fair}.
Devari et al.~\cite{devari2017crowdsourcing} recruited \revise{acquaintances} to \revise{mitigate} privacy concerns in \revise{crowdsourced} delivery.
For food delivery, Wang et al.~\cite{wang2022recommending} presented a recommendation system using \revise{crowdsourced} couriers' preferences. Sun et al.~\cite{sun2019online} examined route recommendations to maximize \revise{crowdsourced couriers'} income. \revise{Chen et al.~\cite{chen2020measuring} found that more than} 14,000 \revise{crowdsourced} taxis \revise{delivered} fewer than 20,000 parcels \revise{per day because of taxi--parcel} matching failures.


\textbf{Cooperative Delivery.}
\revise{The practical} limitations of \revise{UAV delivery}, \revise{courier delivery}, and \revise{crowdsourced delivery motivate} research \revise{on} cooperative delivery. For example, \revise{UAV delivery is constrained by limited battery endurance and flight range~\cite{huang2020method,cai2020500}, courier delivery is labor-intensive and susceptible to capacity pressure as demand grows~\cite{MOHRSSShortage2022,PegadoBardayo2023}, and crowdsourced delivery} faces unstable capacity \revise{because participation is voluntary}~\cite{fatehi2022crowdsourcing}. \revise{Several} hybrid methods \revise{have} emerged: Gao et al.~\revise{\cite{10598004,11261388}} combined UAVs with crowdsourced taxis \revise{by} exploiting taxi trajectories\revise{, while} Pan et al.~\cite{pan2024pioneering} recruited crowdsourced couriers to complement UAVs. \revise{Physical air--ground} cooperation \revise{can further} extend UAV range \revise{by allowing UAVs to use ground transportation networks.} Choudhury et al.~\cite{choudhury2021efficient} used \revise{public-transit} networks to \revise{increase UAV-fleet} range by 450\% \revise{through} task allocation and path planning\revise{, while} Huang et al.~\cite{9151388} \revise{used Dijkstra-based routing to find energy-feasible} paths in \revise{UAV--transit} networks. Pan et al.~\cite{pan2021efficient} \revise{also addressed energy efficiency by recruiting} crowdsourced buses \revise{equipped} with wireless charging \revise{to construct} energy-neutral plans.

In summary, extensive research \revise{has examined UAV} delivery, \revise{courier delivery}, and crowdsourced \revise{delivery}, \revise{but} most studies \revise{consider} these methods separately or \revise{combine only two of them}.
This paper \revise{instead} focuses on integrating \revise{all} three delivery methods \revise{so that their limitations can be offset through coordinated assignment}.



\section{Preliminary}\label{sec.preliminary}

\subsection{{Dataset Description}}\label{sec.dataDescription}

\textbf{Delivery Order Dataset.}
\revise{We use the publicly} released aBeacon dataset~\cite{9524841,10.1145/3372224.3419198}, which \revise{contains} more than 802,000 \revise{historical} parcels delivered \revise{by} over 31,000 couriers in Shanghai during \revise{one} month. Each order \revise{records its} order \revise{and} drop-off times, \revise{as well as restaurant} and \revise{customer} locations. \revise{To protect privacy, the data provider discretized Shanghai into 6,400 square grids of $1\times1\ \mathrm{km}^2$, and locations are represented by grid indices rather than precise GPS coordinates. In this paper, we adopt the same grid system, denoted by $\mathbb{G}=\{g_1,g_2,\ldots,g_{6400}\}$, and divide the 08:00--20:00 study window into 720 one-minute time slots, denoted by $\mathbb{T}=\{t_1,t_2,\ldots,t_{720}\}$.}

\textbf{Trajectory Dataset of GVs.} 
\revise{Because} large-scale \revise{private-car} trajectories \revise{are generally unavailable} due to privacy concerns, we \revise{use} an \revise{open-source} taxi trajectory dataset as \revise{a proxy for ride-hailing-style crowdsourced GVs}. The dataset~\revise{\cite{10.1007/978-3-030-18579-4_45} contains more than} 3.4 billion records \revise{from} over 13,000 taxis in Shanghai during one month. Each record \revise{contains} a timestamp, location, passenger status, and \revise{speed, with a} recording interval \revise{of approximately} 10 seconds. \revise{We map each location to $\mathbb{G}$ and each timestamp within 08:00--20:00 to $\mathbb{T}$, replace direct vehicle identifiers with agent indices, and use only anonymous agent, grid, and time-slot indices in subsequent model inputs and experiments.}


\revise{\textbf{No-Fly-Zone Dataset.}
We obtain the no-fly-zone (NFZ) geofences for Shanghai from DJI~\cite{NFZDJI} and map them onto the grid system $\mathbb{G}$. A grid is labeled as an NFZ if it overlaps any prohibited region. In total, 853 of the 6,400 grids are labeled as NFZs, denoted by $\mathbb{L}_{\mathrm{NFZ}}$.}

\revise{\textbf{Cross-Dataset Harmonization and Experimental Scope.}
The delivery-order and taxi-trajectory datasets originate from different platforms and have nonsynchronized collection periods. Therefore, our experiments constitute a data-driven city-scale simulation rather than an operational replay of a single deployed delivery system. Taxi mobility traces have likewise been used to represent opportunistic delivery capacity in prior urban-delivery studies~\cite{10598004,liu2018foodnet}. We use the aBeacon orders to characterize observed instant-delivery demand and the taxi trajectories as a proxy for ride-hailing-style crowdsourced-GV supply. The two datasets are harmonized only in spatial resolution and within-day temporal resolution: both are mapped to the same $1\times1\ \mathrm{km}^2$ grid system $\mathbb{G}$ and the same 08:00--20:00 one-minute time slots $\mathbb{T}$. This common representation allows candidate parcel--GV assignments to be evaluated according to grid--slot compatibility, but it neither links individual taxis to aBeacon orders nor reconstructs their contemporaneous joint distribution. Consequently, the results should be interpreted as an evaluation of the framework under a ride-hailing-style crowdsourced-GV scenario, rather than as an operational validation using synchronized private-car and delivery-order data.}

\subsection{Motivation}\label{sec.motivation}



Limited \revise{battery} capacity \revise{and} urban flight \revise{approval requirements constrain} the delivery \revise{range and deployable} capacity of UAVs~\cite{9756923,9151388,UAVRegulation}. Pan et al.~\cite{pan2024pioneering} \revise{demonstrated that UAV}--courier \revise{cooperation can substantially reduce} delivery cost \revise{relative to courier-only} delivery. However, crowdsourced courier supply \revise{can be} unstable. \revise{We} therefore \revise{employ} dedicated UAVs and couriers to \revise{provide} stable \revise{baseline} capacity.

\begin{wrapfigure}{r}{4.5cm}  
\vspace{-0.1cm}
    \centering  
    \includegraphics[width=0.25\textwidth]{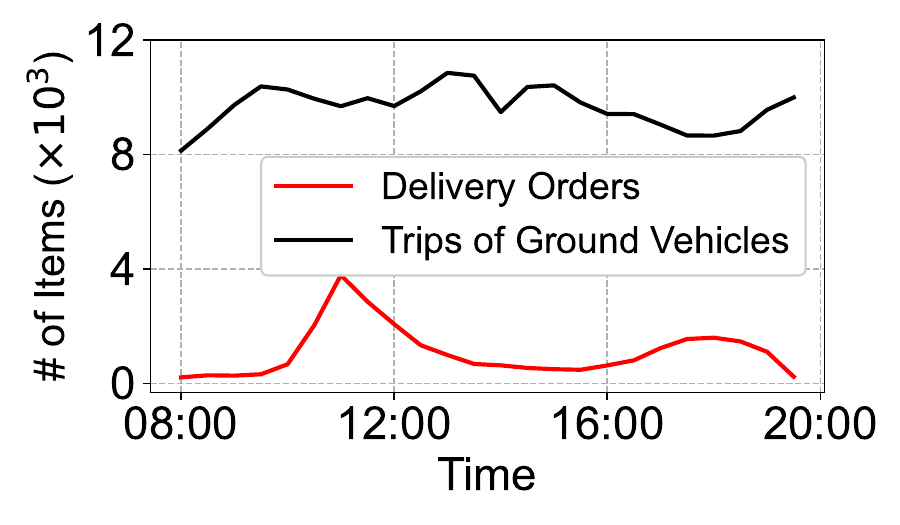}
    \vspace{-0.9cm}
    \caption{\revise{Temporal Distributions} of Delivery Orders and \revise{GV} Trips.}
    \label{fig.temporal_distribution}
    \vspace{-0.4cm}
\end{wrapfigure}

As shown in Fig.~\ref{fig.temporal_distribution}, delivery \revise{demand exhibits pronounced} peaks around lunch (11:00) and dinner (18:00), \revise{while a} large pool of \revise{GV} trips \revise{remains} available \revise{throughout} the \revise{08:00}--20:00 \revise{study window}. \
\revise{Provisioning} dedicated UAVs and couriers \revise{for these demand} peaks would \revise{leave excess} capacity during \revise{off-peak periods. Because} GVs are widely distributed and \revise{can} participate in parcel delivery \revise{on demand}~\cite{liu2018foodnet,chen2020measuring,xu2022drive,7575702}, \revise{they provide elastic supplemental} capacity during \revise{demand surges without} the \revise{same fixed provisioning} cost.

\revise{Accordingly}, {\sf TriDeliver} \revise{uses} dedicated UAVs and couriers for stable \revise{baseline service} and crowdsourced GVs \revise{for elastic peak-demand supplementation}.




\subsection{Framework}\label{sec.framework}
\begin{figure}[t!]
  \centering
  \includegraphics[width=0.49\textwidth]{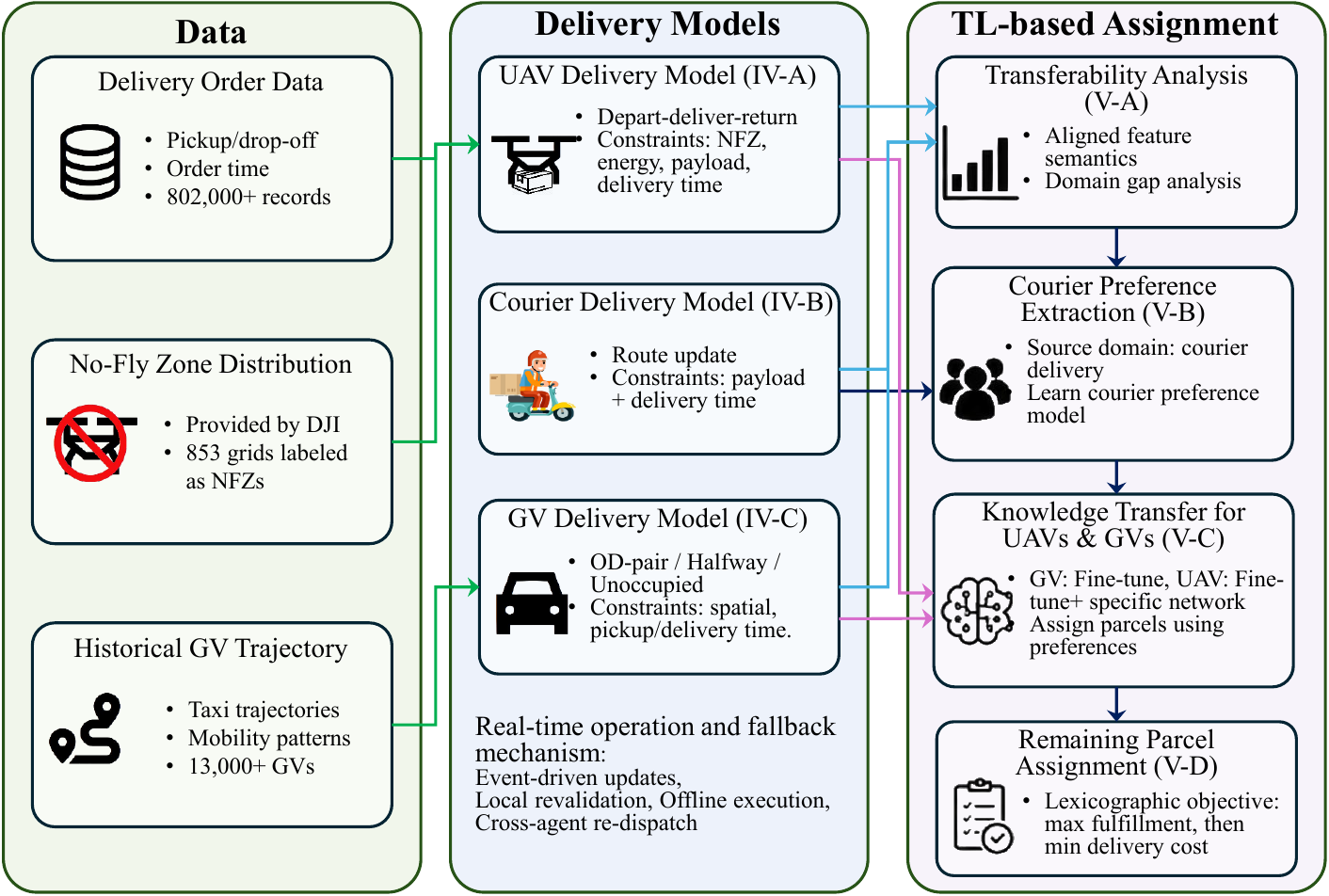}
  \vspace{-0.3cm}
  \caption{Framework of {\sf TriDeliver}.}
  \label{fig.framework}
  \vspace{-0.4cm}
\end{figure}

Fig.~\ref{fig.framework} \revise{organizes} {\sf TriDeliver} \revise{into} three \revise{connected components}: data, delivery models, \revise{and TL-based assignment. Secs}.~\ref{sec.uavModel}, \ref{sec.courierModel}, and \ref{sec.taxiModel} \revise{formulate the agent-specific feasibility constraints and marginal} delivery \revise{costs for} UAVs, couriers, and GVs, respectively. \revise{Sec.~\ref{subsec.transferability} examines knowledge transferability across the three agent domains,} and Sec.~\ref{sec.courierBehaviorExtraction} \revise{learns} courier preferences \revise{from} historical \revise{decisions}. Sec.~\ref{sec.fine-tuning} \revise{transfers this} knowledge to \revise{GVs and} UAVs \revise{to identify feasible highly} preferred \revise{offers, using marginal monetary cost to resolve competition for the same parcel. Finally}, Sec.~\ref{sec.restParcelAssignment} \revise{assigns} the remaining parcels \revise{lexicographically} by first \revise{maximizing fulfillment} and then \revise{minimizing total} delivery \revise{cost}.


\section{Delivery Models}
\label{sec.model}

\revise{This section formulates the agent-specific feasibility constraints and marginal delivery costs used to construct locally verified offers for UAVs, couriers, and GVs. The common communication, acknowledgement, and fallback procedures are detailed in Sec.~\ref{sec.realtimeFallback}.}


\subsection{UAV Delivery Model}\label{sec.uavModel}
\begin{figure}[t!]
    \centering
    \includegraphics[width=0.40\textwidth]{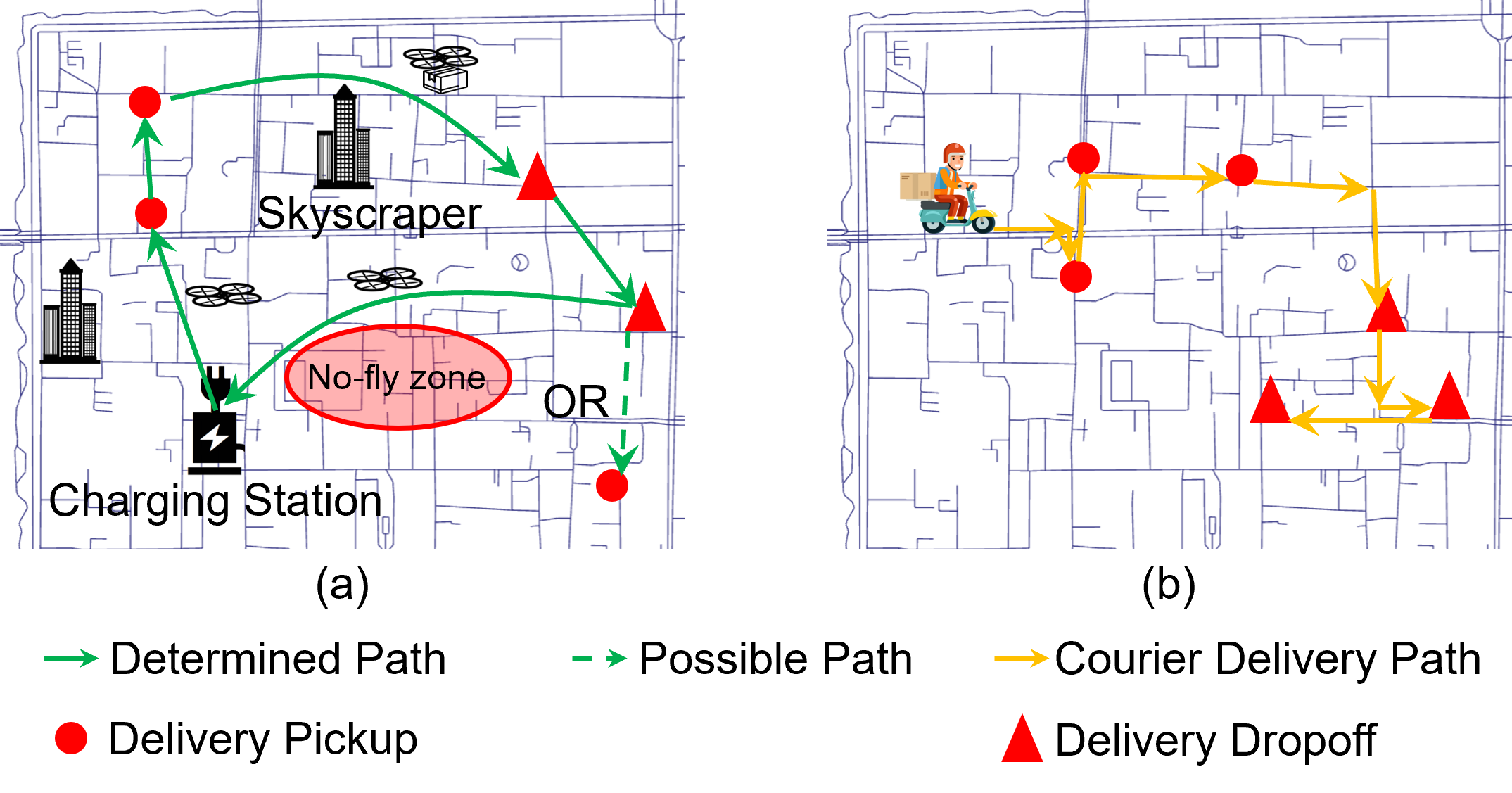}
    \vspace{-0.3cm}
    \caption{Delivery Models of (a) UAVs and (b) couriers.}
    \label{fig.UAVAndCourierModel}
    \vspace{-0.5cm}
\end{figure}


\revise{As shown} in Fig.~\ref{fig.UAVAndCourierModel}(a), a UAV \revise{route comprises three stages}: \textit{departing} from \revise{a} charging \revise{station to a restaurant}, \textit{delivering} the \revise{parcel to a} destination \revise{cabinet}, and \textit{returning} \revise{to a} charging \revise{station}~\cite{xiang2021reusing}. \revise{During} delivery \revise{and return}, the UAV \revise{may evaluate additional feasible} parcels. A sampling-based \revise{planner constructs collision-free} flight \revise{segments that} avoid \revise{NFZs}~\cite{pan2024pioneering}. \revise{Upon returning to a charging station, the UAV swaps its depleted battery for a fully charged one. The station inspects returned batteries, retires those below a predefined capacity threshold, and trickle-charges reusable ones to extend their cycle life~\cite{9756923}.}


Let $\mathbb{U}$ \revise{be the UAV set} and \revise{$\mathbb{P}$ the broadcast parcel set}. $\mathbb{P}(u)$ is the parcels assigned to $u\in\mathbb{U}$. \revise{Its route} is the \revise{ordered} waypoint sequence 
$\Gamma(u,\mathbb{P}(u))=\revise{\bigl(\gamma(u,i)\bigr)_{i=0}^{I}}$\revise{, whose initial} and \revise{terminal} waypoints are charging stations. Each waypoint
$\gamma(u,i)=\langle t(i),l(i),\mathbb{P}(u,i),\mu(i),w(i),E(u,i)\rangle$\revise{
records its} timestamp, location, \revise{processed} parcels, \revise{operation} type, \revise{onboard} weight, and remaining energy, \revise{where }$\mu(i)\in\{1,-1\revise{,0}\}$ denotes \revise{pickup, drop-off, or no parcel operation}. \revise{For each parcel $q\in\mathbb P(u)$, it must be first picked up then delivered, i.e., the pickup and drop-off indices satisfy $0\leq i_q^{\mathrm p}<i_q^{\mathrm d}\leq I$.} 
\revise{$t(i_q^{\mathrm d})$ therefore denotes the drop-off} time \revise{at} the \revise{destination location $l_{\mathrm s}(q)$. Additionally, let $t_{\mathrm o}(q)$} and \revise{$l_{\mathrm o}(q)$} denote the \revise{order} time and \revise{restaurant location} of \revise{$q$}, respectively.
\revise{With $w(q)$ denoting its weight, the onboard load after waypoint $i$ is
$w(i)=\sum_{j=0}^{i}\mu(j)
\sum_{q\in\mathbb{P}(u,j)}w(q),
\quad i=0,\ldots,I.$}



\revise{For consecutive} waypoints, \revise{$\operatorname{seg}_{\mathrm U}(l(i),l(i+1))$ denotes the planned flight segment and
$\operatorname{Dist}_{\mathrm U}(l(i),l(i+1))
=\operatorname{len}(\operatorname{seg}_{\mathrm U}(l(i),l(i+1)))$
its legal flight length. Given UAV speed} $v(u)$, \revise{the next timestamp} is
\revise{$t(i+1)=t(i)+\operatorname{Dist}_{\mathrm U}(l(i),l(i+1))/v(u)$}.



The energy consumed \revise{over} the \revise{planned route} is
\begin{equation}
\footnotesize
\setlength{\abovedisplayskip}{2pt}
\setlength{\belowdisplayskip}{2pt}
\Delta E(u)
=
\sum_{i=0}^{I-1}
\sigma(w(i))
\frac{\operatorname{Dist}_{\mathrm U}(l(i),l(i+1))}{v(u)},
\label{eq.uavEnergyConsumption}
\end{equation}
where $\sigma(w)$ \revise{is} the \revise{payload-dependent energy-consumption} rate. \revise{Following}~\cite{pan2024pioneering,10598004}, \revise{we use}
$\sigma(w)=90.3w+320.9$. \revise{The initial departure and final return segments use the unloaded rate $\sigma(0)$.} \revise{For each} parcel \revise{$q\in\mathbb P(u)$, its} delivery time \revise{is
$t(q)=t(i_q^{\mathrm d})-t_{\mathrm o}(q)$}. The delivery \revise{route $\Gamma(u,\mathbb P(u))$ is feasible only if} the \revise{following four constraints hold}.

\noindent\revise{\textbf{No-Fly-Zone Constraint.}
As a hard constraint, every planned flight segment between consecutive waypoints must satisfy
\begin{equation*}
\footnotesize
\operatorname{seg}_{\mathrm U}(l(i),l(i+1))
\cap\mathbb{L}_{\mathrm{NFZ}}=\emptyset,
\quad i=0,\ldots,I-1.
\end{equation*}
A parcel whose pickup or drop-off lies inside an NFZ is infeasible for UAV delivery and is considered only for couriers or GVs.}

\noindent\revise{\textbf{Payload Constraint.}
The onboard weight must satisfy $0\leq w(i)\leq w_{\max}$ at every waypoint, where $w_{\max}$ is the UAV payload limit.}

\noindent\revise{\textbf{Energy Constraint.}}
Let $E_{\max}$ \revise{be} the \revise{full-battery} energy and \revise{$\alpha\approx10\%$} the \revise{reserve ratio}. \revise{Because segment energy consumption is nonnegative and no in-route recharging is allowed, the reserve throughout the route is enforced by}
\begin{equation}
\footnotesize
E_{\max}-\Delta E(u)\geq\alpha E_{\max}.
\label{eq.uavEnergyConstraint}
\end{equation}
\noindent\revise{\textbf{Delivery-Time Constraint.}}
\revise{Given} the maximum delivery time \revise{$\Delta t$, every} parcel \revise{on} the delivery \revise{route must satisfy}
\begin{equation}
\footnotesize
t(q)\leq\Delta t,
\quad \forall q\in\mathbb{P}(u).
\label{eq.uavDeliveryTimeConstraint}
\end{equation}


When a new parcel \revise{batch} is \revise{announced}, UAV $u$ \revise{evaluates each candidate} parcel \revise{$p$ by tentatively inserting its pickup} and drop-off \revise{operations into the current} delivery \revise{route to construct $\Gamma(u,\mathbb P(u)\cup\{p\})=\bigl(\gamma(u,i)\bigr)_{i=0}^{I'}$, where $I'$ is the terminal waypoint index} of the \revise{tentative route. All waypoint quantities and $\Delta E(u)$ are recomputed on this route}; the \revise{NFZ} constraint is checked \revise{for every consecutive} flight \revise{segment $i=0,\ldots,I'-1$, while the payload} constraint is checked \revise{at every waypoint $i=0,\ldots,I'$. For each existing} parcel \revise{$q\in\mathbb P(u)$, let $\hat t_{\mathrm s}(u,q)$ denote its estimated} drop-off time \revise{on the tentative route, and let $\hat t_{\mathrm s}(u,p)$ denote the estimated} drop-off time of \revise{candidate} parcel \revise{$p$}. The \revise{tentative route must satisfy $\hat t_{\mathrm s}(u,q)-t_{\mathrm o}(q)\leq\Delta t$} for \revise{every $q\in\mathbb P(u)$, together with $\hat t_{\mathrm s}(u,p)-t_{\mathrm o}(p)\leq\Delta t$} for the \revise{candidate} parcel.
\revise{If the tentative route is feasible, the UAV submits a locally verified offer containing $p$, its estimated pickup/drop-off times, remaining-energy margin, marginal delivery cost, and knowledge-induced preference. Selection, acknowledgement, and low-energy fallback follow Sec.~\ref{sec.realtimeFallback}.}



\revise{Let $I$ and $I'$ be} the \revise{terminal} waypoint \revise{indices} of the \revise{current and updated routes}, respectively. \revise{The UAV's marginal monetary} cost \revise{is}
\begin{equation}
\footnotesize
s(u,p)=\revise{0.7}\bigl(t(I')-t(I)\bigr),
\label{eq.uavDeliveryCost}
\end{equation}
\revise{where 0.7 CNY/min is the operating cost per active UAV flight minute, calibrated from public UAV lifecycle-cost estimates and Meituan's reported delivery-time benchmark~\cite{ZTO,UAVCostPerMinute}. The coefficient amortizes equipment and eventual vehicle replacement, routine maintenance, charging-infrastructure operation, battery-related equipment and replacement, energy use, and battery-cycle degradation. Thus, UAV delivery is not treated as cost-free, and its cost is expressed in the same monetary unit as courier and GV costs.}

\revise{Under this calibrated model, UAVs may offer a lower marginal cost when high utilization or route consolidation allows an additional parcel to be served with little extra flight time. This advantage may diminish as maintenance, vehicle- or battery-replacement, or regulatory-compliance costs increase.}


\revise{During flight, the UAV continuously monitors its projected remaining energy. If it is projected to fall below the safety threshold, the UAV locally triggers an emergency update, reports its current state to the control center through the protocol in Sec.~\ref{sec.realtimeFallback}, and is redirected to the nearest available charging station for an immediate battery swap. This mechanism prevents mid-air energy exhaustion and improves real-time robustness.}

\subsection{Courier Delivery Model}\label{sec.courierModel}
Courier delivery also \revise{comprises} departing, delivering, and returning. \revise{Unlike UAVs}, couriers return to \revise{their} stations for rest \revise{only} after \revise{completing} all assigned tasks.
\revise{Let} $\mathbb{C}$ \revise{be} the courier set \revise{and} $\mathbb{P}(c)$ \revise{the} parcels assigned to courier $c\in\mathbb{C}$. The delivery path is \revise{the ordered} waypoint \revise{sequence }$\Gamma(c,\mathbb{P}(c))=\revise{\bigl(\gamma(c,i)\bigr)_{i=0}^{I}}$. Fig.~\ref{fig.UAVAndCourierModel}(b) \revise{illustrates} a courier \revise{who} picks up three parcels at different restaurants and drops them off \revise{sequentially.
Each} waypoint $\gamma(c,i)$ represents a restaurant \revise{pickup or destination-cabinet drop-off and contains} a timestamp $t(i)$, \revise{a} location $l(i)$, \revise{a processed-parcel} set $\mathbb{P}(c,i)$, \revise{an operation} indicator $\mu(i)\in\{1,-1\}$\revise{ for pickup or drop-off}, and the current \revise{parcel count} $n(i)$. \revise{Every parcel has one pickup and one later drop-off on the route. The} payload is limited to $n_{\max}$ \revise{throughout} the \revise{route. Reaching a} waypoint also \revise{triggers an event-driven} status \revise{update} for \revise{online} allocation.
\revise{Let $\operatorname{Dist}_{\mathrm G}(x,y)$ denote} the Manhattan \revise{ground} distance \revise{in kilometers} between \revise{locations $x$} and \revise{$y$; this ground-distance operator is used for both couriers} and \revise{GVs. Using} courier \revise{speed $v(c)$, consecutive waypoint timestamps are updated as $t(i+1)=t(i)+\operatorname{Dist}_{\mathrm G}(l(i),l(i+1))/v(c)$ for $i=0,\ldots,I-1$}.

\revise{Given the delivery process above, the onboard parcel count after processing waypoint $i$} is
\begin{equation}
\footnotesize
\setlength{\abovedisplayskip}{2pt}
\setlength{\belowdisplayskip}{2pt}
n(i) = \revise{\sum_{j=0}^{i}\mu(j)|\mathbb{P}(c,j)|},
\label{eq.carryingNum}
\end{equation}
\revise{where $|\mathbb{P}(c,j)|$} is the number of parcels \revise{processed}
at \revise{waypoint $\gamma(c,j)$}. For \revise{each} parcel \revise{$q\in\mathbb{P}(c)$, $t(q)=t(i_q^{\mathrm d})-t_{\mathrm o}(q)$.
A courier route is feasible only if} the \revise{following constraints hold}.

\noindent\revise{\textbf{Payload Constraint.}}
\revise{The onboard} parcel \revise{count must satisfy}

\begin{equation}
\footnotesize
\setlength{\abovedisplayskip}{1pt}
\setlength{\belowdisplayskip}{2pt}
\revise{0\leq} n(i)\leq n_{\max},
\quad \revise{i=0,\ldots,I.}
\label{eq.payloadCourier}
\end{equation}

\noindent\revise{\textbf{Delivery-Time Constraint.}}
\revise{Every} parcel \revise{on} the courier \revise{route must satisfy}
\begin{equation}
\footnotesize
\setlength{\abovedisplayskip}{2pt}
\setlength{\belowdisplayskip}{1pt}
t(q)\leq\Delta t, \quad \forall q\in\mathbb{P}(c).
\label{eq.deliveryTimeCourier}
\end{equation}

\revise{When} a new parcel \revise{batch} is \revise{announced, each} courier \revise{locally
evaluates the candidate} parcels in \revise{the batch. For each candidate} parcel \revise{$p$},
courier $c$ \revise{tentatively inserts its pickup and drop-off operations into the
current route} to \revise{construct
$\Gamma(c,\mathbb{P}(c)\cup\{p\}) = \bigl(\gamma(c,i)\bigr)_{i=0}^{I'}$,
where $I'$ is the terminal waypoint index} of the \revise{updated
route. For each existing} parcel \revise{$q\in\mathbb P(c)$, let $\hat t_{\mathrm s}(c,q)$ denote its estimated drop-off} time \revise{on the tentative route, and let $\hat t_{\mathrm s}(c,p)$ denote the estimated drop-off} time \revise{of candidate} parcel \revise{$p$}. The courier \revise{then recomputes the waypoint timestamps} and \revise{onboard} parcel \revise{counts, applies} the payload \revise{constraint over $i=0,\ldots,I'$, and requires $\hat t_{\mathrm s}(c,q)-t_{\mathrm o}(q)\leq\Delta t$ for every $q\in\mathbb P(c)$, together with $\hat t_{\mathrm s}(c,p)-t_{\mathrm o}(p)\leq\Delta t$ for} the \revise{candidate} parcel.
\revise{If the updated route is feasible, the courier submits a locally verified
offer containing the candidate parcel, estimated pickup/drop-off times, delivery cost, and preference. The common selection, acknowledgement, and commitment procedures follow
Sec.~\ref{sec.realtimeFallback}.}

\revise{Because detailed rider-compensation rules are not publicly available, we compute the delivery cost reported in each feasible courier offer using the publicly reported distance component of Meituan's fulfillment service fee as a transparent monetary proxy, while omitting context-dependent fee components that cannot be reconstructed from the available data~\cite{MeituanFeeStructure,MeituanFulfillmentFee}.} 
\revise{Define $d_{\mathrm C}(p) =\operatorname{Dist}_{\mathrm G}(l_{\mathrm o}(p),l_{\mathrm s}(p))$ as} the delivery distance \revise{of} parcel $p$. The delivery cost is
\begin{equation}
\footnotesize
\setlength{\abovedisplayskip}{2pt}
\setlength{\belowdisplayskip}{2pt}
s(c,p)
=
\revise{3.15+2\max\{0,d_{\mathrm C}(p)-3\}}.
\label{eq.courierDeliveryCost}
\end{equation}

\subsection{GV Delivery Model}\label{sec.taxiModel}

\begin{figure}[t!]
    \centering
    \includegraphics[width=0.49\textwidth]{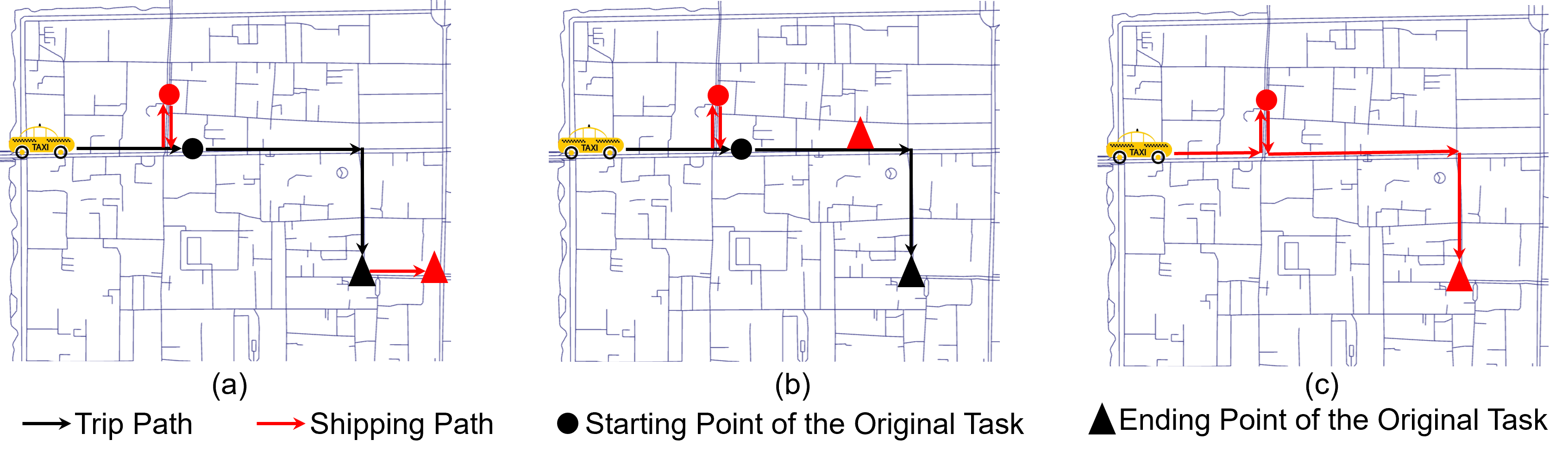}
\vspace{-0.5cm}
    \caption{Delivery Cases of GVs. (a) \textit{OD-pair Delivery}; (b) \textit{Halfway Delivery}; (c) \textit{Unoccupied Delivery}.}
    \label{fig.taxiDeliveryModel}
    \vspace{-0.5cm}
\end{figure}
\revise{Let $\mathbb{B}$ be the GV set and consider a candidate agent--parcel pair $(b,p)$ with $b\in\mathbb{B}$.} \revise{We consider} three \revise{GV} delivery \revise{modes according to} the \revise{parcel locations} and
the \revise{GV's} original \revise{task}: \textit{\revise{Origin--Destination} (OD)-pair Delivery},
\textit{Halfway Delivery}, and \textit{Unoccupied Delivery}, \revise{as} shown in
Fig.~\ref{fig.taxiDeliveryModel}~\cite{7575702,LI201431,liu2018foodnet}.
\revise{Each GV serves at most one parcel at a time. For
\textit{OD-pair Delivery} and \textit{Halfway Delivery}, an original task is
an accepted but not-yet-started passenger order, so parcel pickup occurs
before the passenger trip begins. The estimated pickup and drop-off times $\hat t_{\mathrm{pu}}(b,p)$ and $\hat t_{\mathrm s}(b,p)$ under a candidate delivery mode are obtained by accumulating the travel times of the corresponding ground-route
segments, where the travel time from $x$ to $y$ is
$\operatorname{Dist}_{\mathrm G}(x,y)/v(b)$.}

In \textit{OD-pair Delivery}~\cite{7575702}, GV $b$ picks up parcel $p$,
\revise{completes its} original \revise{passenger trip without performing} a parcel \revise{operation,
and then delivers} $p$ \revise{from} the \revise{original-trip} destination to its destination
cabinet.

\noindent\textbf{Spatial \revise{Constraint.}}
\revise{Let $d_{\mathrm{pu}}(b,p)$ be the additional} distance \revise{required} to pick up
$p$ \revise{before} the \revise{passenger trip}.
\revise{Let $l_{\mathrm{drop}}^{\mathrm{ori}}(b)$ be the original-trip
destination and define
$d_{\mathrm{do}}(b,p)
=\operatorname{Dist}_{\mathrm G}
(l_{\mathrm{drop}}^{\mathrm{ori}}(b),l_{\mathrm s}(p))$
as the post-trip delivery distance.}
\textit{\revise{OD-pair Delivery}} is \revise{feasible only if
}$\revise{d_{\mathrm{pu}}(b,p)}\leq d_{\max}$ \revise{and
$d_{\mathrm{do}}(b,p)\leq d_{\max}^{\mathrm{do}}$.}

\noindent\revise{\textbf{Pickup-Time Constraint.}}
\revise{Because} parcel pickup \revise{occurs before} the original \revise{passenger trip begins, we bound the pre-trip} pickup \revise{interval to} limit \revise{delivery-induced delay to that} task: \revise{$\hat t_{\mathrm{pu}}(b,p)-t_{\mathrm o}(p)\leq\Delta t_{\mathrm{pu}}$, where} $\Delta t_{\mathrm{pu}}$ \revise{is} the \revise{maximum allowable interval between} parcel \revise{order and} pickup \revise{before} the \revise{passenger trip begins}.

\revise{In} \textit{Halfway Delivery}~\cite{LI201431}, the \revise{GV picks up $p$ before the passenger trip} and \revise{drops it} off when \revise{its planned route passes near} the parcel \revise{destination. This mode must satisfy} the \revise{pickup-distance} and \revise{pickup-time constraints above, together with }$d'(b,p)\leq\Delta d$\revise{, where} $d'(b,p)$ is \revise{the total additional ground-route distance caused by} the drop-off \revise{micro-detour}.

In \textit{Unoccupied Delivery}~\cite{liu2018foodnet,10598004}, \revise{an idle GV performs} the \revise{complete pickup-and-delivery} trip. \revise{In all} three \revise{modes}, the \revise{estimated drop-off} time \revise{must satisfy $\hat t_{\mathrm s}(b,p)-t_{\mathrm o}(p)\leq\Delta t$}.

\revise{The three modes are evaluated locally by the vehicle-side module. For a GV with an original task, the passenger service order provides the destination, and the local navigation system provides the planned route. \textit{Halfway Delivery} is offered only when the GV locally confirms that this route passes sufficiently close to the parcel destination. The GV reports only the feasible mode, estimated pickup and drop-off times, detour distance, delivery cost, and preference, without disclosing its passenger destination or complete future route. It updates or withdraws the offer if the route or feasibility changes before selection. Before commitment, a selected GV rechecks route overlap, detour tolerance, and timing constraints. If this check fails, the
GV rejects the assignment and the parcel is reconsidered using the remaining
offers; it remains unassigned if none is feasible. 
The common acknowledgement
and fallback procedures, including re-dispatch of confirmed but unpicked parcels and recovery re-dispatch of picked but undelivered parcels after agent unavailability, are detailed in Sec.~\ref{sec.realtimeFallback}.}

Parcel delivery \revise{may interfere with a GV's} original \revise{task, so participating}
drivers \revise{must be compensated. Let} $d(b,p)$ \revise{denote the mode-dependent} detour
\revise{distance: it equals the pickup} detour \revise{plus the post-trip} delivery \revise{distance}
for \textit{OD-pair Delivery}, \revise{the pickup} detour for
\textit{\revise{Halfway} Delivery}, and the \revise{complete pickup-and-delivery distance for}
\textit{Unoccupied Delivery}. \revise{The term} $d'(b,p)$ \revise{denotes the}
\textit{\revise{Halfway} Delivery} \revise{drop-off micro-detour and is zero} in the other
\revise{modes. Using} the Shanghai taxi unit \revise{fare} of 2.7 CNY/\revise{km}~\cite{ShanghaiTaxiFee},
the delivery cost is

\begin{equation}
\footnotesize
s(b,p)
=
\revise{\lambda}\times2.7\times\bigl(d(b,p)+d'(b,p)\bigr).
\label{eq.gvDeliveryCost}
\end{equation}

\revise{The coefficient is $\lambda=2$ for} GVs \revise{with an} original \revise{passenger task:
one} unit \revise{pays} for the delivery \revise{distance} and the other \revise{compensates} for
\revise{interference with that task}. For \textit{Unoccupied Delivery}, \revise{no} original
\revise{task is affected, so $\lambda=1$}.

\noindent\revise{\textbf{Commercial Pricing Interpretation.}
The costs $s(u,p)$, $s(c,p)$, and $s(b,p)$ are internal marginal costs used
for assignment, rather than customer-facing prices. A delivery company may
set customer prices according to distance, urgency, peak-hour conditions,
platform overhead, and its target profit margin. For a fixed set of delivered
parcels and exogenously determined customer prices, minimizing internal
fulfillment cost increases the assignment-level operating margin. Thus,
{\sf TriDeliver} does not prescribe a particular end-user pricing rule.}

\subsection{\revise{Real-Time Operation and Fallback Mechanism}}
\label{sec.realtimeFallback}

\revise{To cope with network congestion and intermittent connectivity, the system uses an event-driven, metadata-only protocol rather than streaming raw trajectories or complete future routes. Messages are exchanged only when a parcel batch is announced, a scheduled waypoint is reached, a route change requires an offer update or withdrawal, or an abnormal state such as low battery or prolonged disconnection is detected. Common fields include the agent ID, candidate parcel ID, feasible delivery mode, estimated pickup/drop-off times, marginal delivery cost, and preference. Agent-specific fields are the current load and route availability for couriers, payload and remaining-energy margin for UAVs, and trip availability and estimated detour for GVs; a GV does not disclose its complete destination or future route. For each locally feasible agent--parcel pair, the agent evaluates the preference on its agent-side module and includes the score in its submitted offer. Notably, the control center neither re-evaluates the preference level nor receives the complete agent route.}

\revise{Based on the uploaded agent-side offers, the control center assigns parcels and sends assignment requests. A selected agent acknowledges only if its offer remains current; otherwise, it returns a compact rejection. A selected GV additionally rechecks route overlap, detour tolerance, and timing constraints as described in Sec.~\ref{sec.taxiModel}. Only acknowledged assignments are committed; a rejected agent--parcel pair is removed for reassignment. 
By replacing continuous reporting with event-driven transmission of compact, assignment-relevant metadata, this protocol reduces both transmission frequency and per-message data volume, thereby lowering bandwidth demand and helping to mitigate congestion-induced communication delay. The resulting lightweight offer/acknowledgement exchanges can be supported by 5G-capable low-latency modules~\cite{campion2018uav,9248613,7876982}.}

\revise{The effect of crowded roadways is reflected through the agent-specific ground speeds used to estimate courier and GV pickup/drop-off times. Lower ground speeds increase these estimates and may invalidate a ground-agent offer under the delivery-time constraints. When a legal UAV offer exists, part of this traffic-sensitive workload can be shifted from ground agents to UAVs. Conversely, if an NFZ removes the UAV alternative and no feasible ground offer remains, the parcel is left unassigned rather than being forced into an infeasible route. Thus, heterogeneous cooperation can mitigate part of the capacity loss when at least one feasible alternative offer exists.}

\revise{During a temporary disconnection, an agent may complete a parcel already picked up using cached instructions and synchronizes after reconnection, whereas a confirmed but unpicked parcel is held until the link is restored. Versioning rejects stale instructions and prevents concurrent offline execution and central re-dispatch.}


\revise{An agent is marked unavailable if it misses scheduled waypoint reports, remains disconnected for a prolonged period, or reports that it can no longer satisfy the delivery constraints. All of its offers are withdrawn. A confirmed but unpicked parcel is revoked and returned to the ordinary re-dispatch process. A temporary communication outage alone does not trigger reassignment of a parcel already in transit; recovery begins only after the agent is confirmed to be unable to complete delivery. For each picked but undelivered parcel $p$, let $l_{\mathrm f}(p)$ and $t_{\mathrm f}(p)$ denote the confirmed accessible failure/handoff location and the time at which the parcel becomes available for custody transfer, respectively. The parcel is then converted into a recovery task with effective pickup location $\tilde l_{\mathrm o}(p)=l_{\mathrm f}(p)$ and recovery release time $\tilde t_{\mathrm o}(p)=t_{\mathrm f}(p)$, while its original destination and other parcel attributes remain unchanged. For all subsequent route, feature, marginal-cost, preference-score, and feasibility calculations, these effective recovery fields replace the original pickup fields. A replacement agent $a'$ must therefore satisfy $\hat t_{\mathrm s}(a',p)-\tilde t_{\mathrm o}(p)\leq\Delta t$. The original order time $t_{\mathrm o}(p)$ is nevertheless retained for end-to-end service reporting. The failed assignment and all stale offers are invalidated, and the remaining UAVs, couriers, and GVs generate fresh locally verified offers from the recovery state. A versioned custody-transfer record allows the replacement assignment to be committed only after handoff availability is confirmed, thereby preventing duplicate custody. If no feasible offer is currently available, the parcel remains pending at the recovery location for a later allocation cycle rather than being discarded. The same recovery procedure applies when a UAV failure reduces flight capacity. The current evaluation does not inject stochastic physical failures; this paragraph specifies the system-level recovery behavior rather than an experimentally evaluated performance result.}

\section{Transfer Learning-based Parcel Assignment}
\label{sec.TLAssignment}


In this section, we \revise{first analyze the transferability of courier-side delivery knowledge to GVs and UAVs. We then extract the courier preference model and adapt it to the two target domains through fine-tuning and agent-specific components. Finally, parcels not committed in the preference-based stage are handled by a residual assignment model that lexicographically prioritizes fulfillment over monetary cost.}

\subsection{\revise{Transferability Analysis}}
\label{subsec.transferability}

\revise{Before presenting the detailed TL-based assignment strategy, we first clarify why courier-side delivery knowledge can be transferred to UAVs and GVs despite their heterogeneous delivery patterns. The proposed transfer does not aim to directly transfer couriers' physical movement patterns, such as ground routes or riding behaviors. Instead, the transferred object is the high-level parcel-assignment decision function learned from mature courier delivery behaviors. Such a function captures shared, decision-relevant scheduling cues in instant delivery, including spatio-temporal demand regularities, parcel urgency, the tradeoff between delivery time and delivery cost, detour burden, and the influence of agent-side resource states on parcel-selection preferences. For example, courier behavior can provide an initial cue that urgent parcels should be matched with agents capable of timely service and that assignments requiring large detours are less attractive; target-domain fine-tuning then adapts these cues to UAV-specific flight and endurance constraints and GV-specific passenger-trip detours.}

\revise{Although couriers, UAVs, and GVs operate under different physical constraints, their preference feature vectors contain the same categories of assignment information: order time and location, detour, speed, delivery distance and cost, current load/task state, and a time-related state. The variable used for each item follows the corresponding delivery model. For example, the courier parcel-count feature $n(c)$ is replaced by the UAV carrying weight $w(u)$, while the courier-side ground distance is replaced by the legal UAV flight distance. The agent-specific variables are defined below, and each preference model learns how they contribute to its preference score from the corresponding domain data. This common feature organization allows the target models to be initialized from the courier model while retaining variables that reflect their own physical constraints; it does not assume that the three feature distributions are identical.}


\revise{To quantify the remaining domain discrepancies, we compute MMD over the eight aligned scalar features: location, order time, detour, speed, delivery distance, delivery cost, current load/task state, and the time-related state. Each feature is standardized using the mean and population standard deviation computed from all available courier, GV, and UAV records pooled together. In each of ten repetitions (random seed 2026), every comparison samples 5,000 records per domain without replacement, with two disjoint courier subsets used for the Courier--Courier reference. We use the biased MMD$^2$ estimator with an RBF kernel whose bandwidth is selected for each sampled comparison by the median heuristic using at most 2,000 pooled records. Table~\ref{tab.mmd_domain_gap} reports the mean MMD$^2$ and its square root. Both source--target MMD values are far above the Courier--Courier reference, indicating discrepancies much larger than the observed intra-domain variation; the Courier--UAV MMD is slightly higher than the Courier--GV MMD.}

\begin{table}[t]
\centering
\caption{\revise{MMD-based Quantification of Source-Target Domain Gaps.}}
\label{tab.mmd_domain_gap}
\footnotesize
\begin{tabular}{lccc}
\hline
\revise{Comparison} & \revise{Samples per domain} & \revise{MMD$^2$} & \revise{MMD} \\
\hline
\revise{Courier-Courier} & \revise{5,000} & \revise{0.0003} & \revise{0.0177} \\
\revise{Courier-GV}      & \revise{5,000} & \revise{0.2893} & \revise{0.5379} \\
\revise{Courier-UAV}     & \revise{5,000} & \revise{0.3180} & \revise{0.5639} \\
\hline
\end{tabular}
\end{table}

\revise{Both source--target MMD values are far above the intra-domain reference, motivating target-domain adaptation for both agent types. The slightly lower Courier-GV MMD is consistent with the fact that couriers and GVs both operate on road networks and face ground travel-time, congestion, and detour constraints; we therefore adapt the courier preference model to GVs through fine-tuning. The UAV domain additionally involves qualitatively different legal flight paths, no-fly zones, payload-dependent energy consumption, and endurance constraints, and its MMD is higher. We therefore append a Specific Network after fine-tuning to capture this UAV-specific structure. This asymmetric design reuses courier-derived assignment knowledge while adapting it to the two heterogeneous agent types.}

\subsection{Courier Preference Extraction and Representation}\label{sec.courierBehaviorExtraction}
\revise{We represent each courier--parcel decision using spatial, mobility, monetary, workload, and timing features.} 
\revise{For the $i$-th labeled courier--parcel record, $(c_i,p_i)$ identifies
the courier and parcel involved. Based on this pair and the delivery
state at the decision time, we construct the feature vector
$x_i^{\mathrm C}$ (with ${N_{\mathrm C}}$ courier--parcel pairs) as follows:}
\begin{equation}
    \begin{split}
    \revise{{x_i^{\mathrm C}}}=&[{t_{\mathrm o}(\revise{p_i})},{l_{\mathrm o}(\revise{p_i})},d(\revise{c_i},\revise{p_i}),v(\revise{c_i}),\\
    &{\operatorname{Dist}_{\revise{\mathrm G}}(l_{\mathrm o}(\revise{p_i}),l_{\mathrm s}(\revise{p_i}))},s(\revise{c_i},\revise{p_i}),n(\revise{c_i}),\\
    &\revise{t_{\mathrm{re}}(c_i,p_i)}],\revise{\quad i=1,\ldots,{N_{\mathrm C}}},\\
    {\mathcal{X}_{\mathrm C}}=&\revise{({x_i^{\mathrm C}})_{i=1}^{{N_{\mathrm C}}}}.
    \end{split}
    \label{eq.courierFeature}
\end{equation}
\revise{Here, $n(c_i)$ is the number of parcels currently carried by courier $c_i$, and $d(c_i,p_i)$ is the marginal route-insertion distance. 
Let $\mathbb{P}_{\mathrm{tr}}(c_i)$ denote the set of parcels
currently carried by $c_i$ and not yet delivered.
The courier temporal margin is}
\begin{equation*}
\footnotesize
\revise{t_{\mathrm{re}}(c_i,p_i)
=
\min_{q\in\mathbb{P}_{\mathrm{tr}}(c_i)\cup\{p_i\}}
\left[
\Delta t-
\left(
\hat t_{\mathrm{s}}(c_i,q)-t_{\mathrm{o}}(q)
\right)
\right].}
\end{equation*}
\revise{Here, $\hat t_{\mathrm{s}}(c_i,q)$ is the estimated drop-off time of
parcel $q$ on the tentative route after inserting candidate parcel $p_i$.
Thus, $t_{\mathrm{re}}(c_i,p_i)$ is the minimum remaining
delivery-deadline slack among all in-transit parcels and the candidate
parcel. A larger value indicates greater temporal flexibility, whereas a
negative value indicates that at least one parcel would violate the
delivery-time constraint.
At inference time, $x(c,p)$ denotes the feature vector constructed by the same mapping for any candidate pair $(c,p)$.}

\begin{figure*}[t!]
    \centering
    \includegraphics[width=0.7\textwidth]{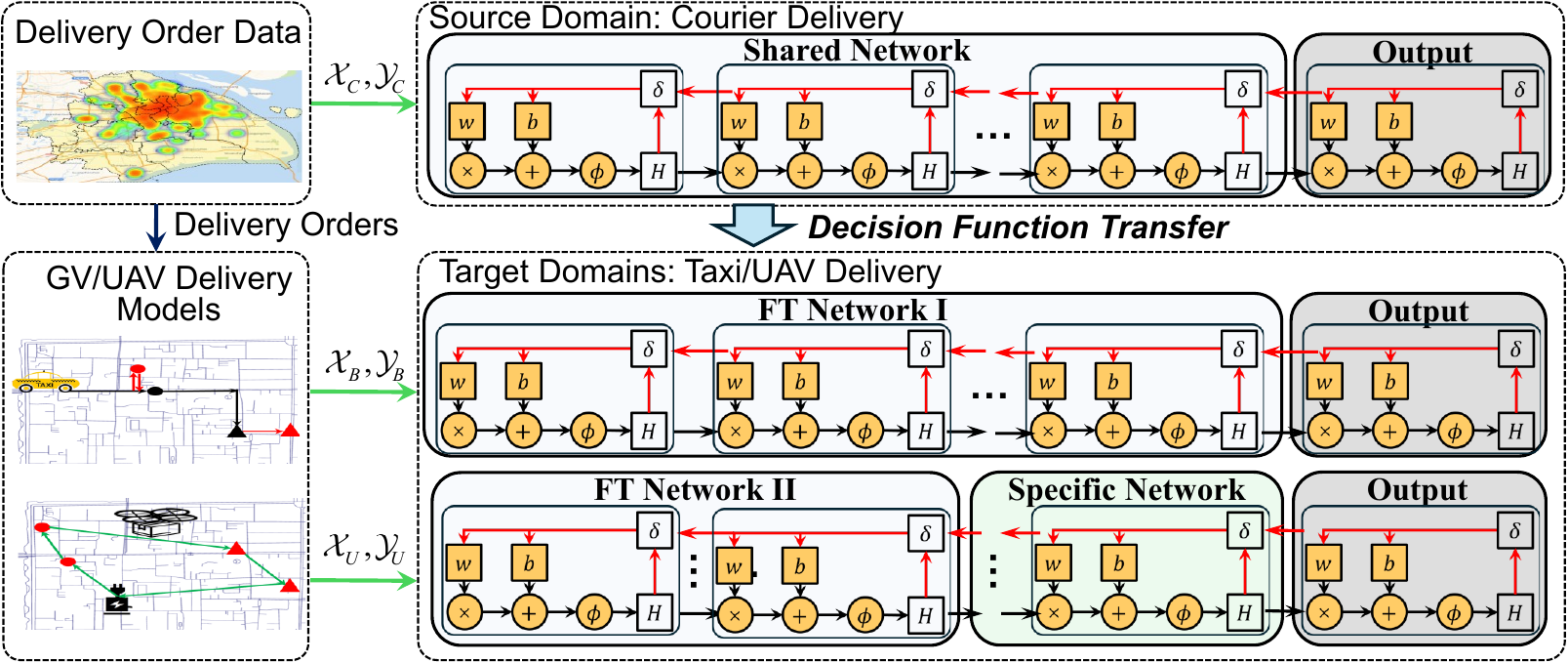}
    \caption{Transfer and Fine-tuning of Decision Functions.}
    \label{fig.TL}
\end{figure*}

Let ${\mathcal{Y}_{\mathrm C}}$ \revise{be the matched label sequence, where 1 and 0 denote positive and negative preference labels, respectively:}
\begin{equation}
    {\mathcal{Y}_{\mathrm C}}=\revise{({y_i^{\mathrm C}})_{i=1}^{{N_{\mathrm C}}}},
    \qquad \revise{{y_i^{\mathrm C}}=}y(\revise{c_i},\revise{p_i})\in\{0,1\}.
    \label{eq.label_space_courier}
\end{equation}
\revise{An observed courier--parcel pair is labeled positive if the courier delivered the parcel. Negative records are constructed from parcels not delivered by that courier.
The resulting negative pool is randomly subsampled to keep the two classes approximately balanced.}
The Shared Network in Fig.~\ref{fig.TL} \revise{learns a courier} decision function \revise{${f_{\mathrm C}}$. Its output
$\rho(c,p)={f_{\mathrm C}}(x(c,p))\in[0,1]$ is the sigmoid-bounded preference score for courier--parcel pair $(c,p)$.}
\revise{Let ${H_{\mathrm C}^{(k)}}$ be} the input \revise{to layer $k$ of an} $M$\revise{-layer network. The forward propagation is}
\begin{equation}
    {H_{\mathrm C}^{\revise{(k+1)}}}=\revise{\psi_k}\!\left({H_{\mathrm C}^{\revise{(k)}}}{W_{\mathrm C}^{\revise{(k)}}}+{\beta_{\mathrm C}^{\revise{(k)}}}\right),
    \ \revise{k=0,\ldots,M-1},
\end{equation}
\revise{where ${W_{\mathrm C}^{(k)}}$ and ${\beta_{\mathrm C}^{(k)}}$ are} the weight \revise{matrix and bias vector.} Hidden layers use ReLU\revise{, and} the output layer \revise{uses a} sigmoid \revise{activation. For sample $i$, the predicted preference score is $\hat{y}_i^{\mathrm C}=f_{\mathrm C}(x_i^{\mathrm C})$. We train the network with binary cross-entropy between $\hat{y}_i^{\mathrm C}$ and the training label $y_i^{\mathrm C}$:}
\begin{equation}
\setlength{\abovedisplayskip}{2pt}
\setlength{\belowdisplayskip}{2pt}
    L = -\frac{1}{{N_{\mathrm C}}}\sum_{i=1}^{{N_{\mathrm C}}}
    \left[{y_i^{\revise{\mathrm C}}}\log({\hat{y}_i^{\revise{\mathrm C}}})+(1-{y_i^{\revise{\mathrm C}}})\log(1-{\hat{y}_i^{\revise{\mathrm C}}})\right].
    \label{eq.courier_loss}
\end{equation}
\revise{We minimize this loss with} Adam\revise{, which combines} adaptive learning rates \revise{with} momentum-based \revise{updates~\cite{kingma2014adam}.}

\subsection{Knowledge Transfer and Fine-tuning for GVs and UAVs}\label{sec.fine-tuning}
\revise{Following the transferability analysis above, the transfer is performed over semantically aligned feature spaces rather than over raw physical trajectories.}

\revise{Because} couriers, GVs, and UAVs \revise{follow} different delivery \revise{processes, their} feature spaces \revise{use different physical variables with aligned decision roles.} \revise{Let $N_{\mathrm B}$ denote the number of labeled GV--parcel records. For record $i=1,\ldots,N_{\mathrm B}$, $(b_i,p_i)$ identifies the GV and parcel involved. Based on this pair and the GV state at the candidate-evaluation time, we construct $x_i^{\mathrm B}$.} \revise{In this vector, courier parcel count} is replaced \revise{by the binary original-task commitment indicator $n(b_i)$, where 1 denotes an accepted but not-yet-started original passenger task and 0 denotes an idle GV:}
\begin{equation}
    \begin{split}
    \revise{{x_i^{\mathrm B}}}=&[{t_{\mathrm o}(\revise{p_i})},{l_{\mathrm o}(\revise{p_i})},d(\revise{b_i},{\revise{p_i}}),v(\revise{b_i}),\\
    &{\operatorname{Dist}_{\revise{\mathrm G}}(l_{\mathrm o}(\revise{p_i}),l_{\mathrm s}(\revise{p_i}))},s(\revise{b_i},{\revise{p_i}}),n(\revise{b_i}),\\
    &\revise{t_{\mathrm{op}}(b_i)}],\revise{\quad i=1,\ldots,{N_{\mathrm B}}},\\
    {\mathcal{X}_{\mathrm B}}=&\revise{({x_i^{\mathrm B}})_{i=1}^{{N_{\mathrm B}}}},
    \end{split}
    \label{eq.taxiFeature}
\end{equation}
\revise{Here, $t_{\mathrm{op}}(b)$ represents} the scheduled \revise{original-trip duration for a GV with an} original task \revise{and uses a fixed value for an idle GV:}
\begin{equation*}
\revise{t_{\mathrm{op}}(b)=
\begin{cases}
\left(t_{\mathrm{drop}}^{\mathrm{ori}}(b)-t_{\mathrm{pick}}^{\mathrm{ori}}(b)\right),
& n(b)=1,\\
\Delta t, & n(b)=0,
\end{cases}}
\end{equation*}
\revise{where $t_{\mathrm{pick}}^{\mathrm{ori}}(b)$ and $t_{\mathrm{drop}}^{\mathrm{ori}}(b)$ are the scheduled pickup and drop-off times of GV $b$'s original passenger trip. For a GV with an accepted but not-yet-started original task, this feature is the full scheduled trip duration; for an idle GV, it is set to the delivery-time limit $\Delta t$. A longer trip may provide more route-overlap opportunities for Halfway Delivery but postpone the post-trip parcel leg in OD-pair Delivery. Feasibility is checked locally using route overlap, pickup and post-trip distances, and pickup time.} \revise{Because GVs travel on streets,} the Manhattan distance \revise{${\operatorname{Dist}_{\mathrm G}(l_{\mathrm o}(p),l_{\mathrm s}(p))}$} remains in ${\mathcal{X}_{\mathrm B}}$.
\revise{Let $N_{\mathrm U}$ denote the number of labeled UAV--parcel records. For record $i=1,\ldots,N_{\mathrm U}$, $(u_i,p_i)$ identifies the UAV and parcel involved. Based on this pair and the UAV state at the candidate-evaluation time, we construct $x_i^{\mathrm U}$.} \revise{In this vector, courier parcel count is replaced by onboard weight $w(u_i)$ because UAV feasibility depends on payload weight}
\begin{equation}
    \begin{split}
    \revise{{x_i^{\mathrm U}}}=&[{t_{\mathrm o}(\revise{p_i})},{l_{\mathrm o}(\revise{p_i})},d(\revise{u_i},{\revise{p_i}}),v(\revise{u_i}),\\
    &{\operatorname{Dist}_{\mathrm U}(l_{\mathrm o}(\revise{p_i}),l_{\mathrm s}(\revise{p_i}))},s(\revise{u_i},{\revise{p_i}}),w(\revise{u_i}),\\
    &\revise{t_{\mathrm{re}}(u_i,p_i)}],\revise{\quad i=1,\ldots,{N_{\mathrm U}}},\\
    {\mathcal{X}_{\mathrm U}}=&\revise{({x_i^{\mathrm U}})_{i=1}^{{N_{\mathrm U}}}},
    \end{split}
    \label{eq.uavFeature}
\end{equation}
\revise{Here, $d(u_i,{p_i})$ denotes $u$'s marginal route-insertion distance.}
\revise{The feature $t_{\mathrm{re}}(u_i,p_i)$ is the residual flight endurance computed from the UAV's energy at the candidate pickup insertion point and its consumption rate under the current payload. An idle UAV is assigned its full-energy endurance, so a larger value indicates more remaining flight capacity.} \revise{The corresponding binary label sequences are ${\mathcal{Y}_{\mathrm B}}=({y_i^{\mathrm B}})_{i=1}^{{N_{\mathrm B}}}$ and ${\mathcal{Y}_{\mathrm U}}=({y_i^{\mathrm U}})_{i=1}^{{N_{\mathrm U}}}$, where $y_i^{\mathrm B}=y(b_i,p_i)$ and $y_i^{\mathrm U}=y(u_i,p_i)$. Their values 1 and 0 have the same meanings as in Eq.~\eqref{eq.label_space_courier}.}
\revise{At inference time, $x(b,p)$ and $x(u,p)$ denote features constructed by the corresponding mappings above.}
\revise{To construct the target-domain training records}, we simulate cost-oriented \revise{GV} and UAV delivery using the heuristic algorithms \revise{of} Gao et al.~\cite{10598004} \revise{and historical demand and mobility data.} \revise{For each parcel, the applicable route, timing, capacity, payload, energy, and original-task constraints are checked before labels are generated. The feasible candidate selected by the heuristic is labeled 1, and the lowest-ranked feasible candidate under the same heuristic is labeled 0 when available. Thus, these labels distinguish the heuristic choices within the feasible candidate set rather than encode feasibility itself. Because large-scale deployed UAV/GV delivery logs are not publicly available, the generated records serve as cold-start training data and can later be supplemented with real deployment logs when available.}
\revise{The GV decision function ${f_{\mathrm B}}$, represented by} FT Network I in Fig.~\ref{fig.TL}\revise{, is initialized from the trained courier network and fine-tuned on $({\mathcal{X}_{\mathrm B}},{\mathcal{Y}_{\mathrm B}})$. At optimization iteration $r$, each layer $k$ is updated by standard gradient descent:}
\begin{equation}
\setlength{\abovedisplayskip}{2pt}
\setlength{\belowdisplayskip}{1pt}
    W_k^{\revise{(r+1)}}=W_k^{\revise{(r)}}-\eta
    \frac{\partial L_{\mathrm{pref}}}{\partial W_k^{\revise{(r)}}},
    \label{eq.weightUpdates}
\end{equation}
\begin{equation}
\setlength{\abovedisplayskip}{0pt}
\setlength{\belowdisplayskip}{2pt}
    \beta_k^{\revise{(r+1)}}=\beta_k^{\revise{(r)}}-\eta
    \frac{\partial L_{\mathrm{pref}}}{\partial \beta_k^{\revise{(r)}}},
    \label{eq.offsetUpdates}
\end{equation}
where $\eta$ \revise{is} the learning rate \revise{and $L_{\mathrm{pref}}$ is the binary cross-entropy loss on the target-domain fine-tuning records. The gradients are computed by backpropagation.}

\revise{Unlike} couriers and GVs, UAVs \revise{are unaffected by ground traffic but face aerial route, payload, and endurance constraints}. To capture \revise{this larger residual domain gap, the UAV branch appends a} Specific Network \revise{to the fine-tuned shared network,} as shown in Fig.~\ref{fig.TL}\revise{. The shared component is initialized from the courier model and fine-tuned on $({\mathcal{X}_{\mathrm U}},{\mathcal{Y}_{\mathrm U}})$ using Eqs.~\eqref{eq.weightUpdates} and~\eqref{eq.offsetUpdates}; the} Specific Network \revise{then learns the remaining UAV-specific structure.}

\revise{The three} decision functions \revise{combine courier-derived knowledge with agent-specific} delivery patterns\revise{. Let $\rho(c,p)={f_{\mathrm C}}(x(c,p))$, $\rho(b,p)={f_{\mathrm B}}(x(b,p))$, and $\rho(u,p)={f_{\mathrm U}}(x(u,p))$ be the corresponding preference scores.} \revise{At assignment time, each agent-side module first checks the route, capacity, timing, energy, and other operational constraints applicable to a candidate parcel. For every agent--parcel pair $(a,p)$ that passes this check, the module constructs $x(a,p)$, evaluates the corresponding deployed preference function, and includes $\rho(a,p)$ in the offer sent to the control center. The control center uses the reported preference score and marginal cost below without reconstructing $x(a,p)$ or receiving the complete local route. Therefore, every high-preference offer is feasible by construction, whereas a feasible offer may not exceed the corresponding preference threshold. For each pair $(a,p)$, the local module maintains at most one active offer; a newer offer version supersedes the previous one. The record binds the locally retained tentative plan and its selected feasible mode and stores the estimated pickup/drop-off times, marginal cost, and preference score. For parcel $p$, define the set of agents with active high-preference offers as}
\begin{equation}
\setlength{\abovedisplayskip}{1pt}
\setlength{\belowdisplayskip}{1pt}
    \begin{aligned}
    \revise{\mathcal{F}(p)={}}&
    \revise{\{c\in\mathbb{C}\mid (c,p)\text{ feasible},\ }\rho(c,p)\revise{\geq}{\epsilon_{\mathrm C}}\revise{\}}\\
    &\revise{\cup\{b\in\mathbb{B}\mid (b,p)\text{ feasible},\ }\rho(b,p)\revise{\geq}{\epsilon_{\mathrm B}}\revise{\}}\\
    &\revise{\cup\{u\in\mathbb{U}\mid (u,p)\text{ feasible},\ }\rho(u,p)\revise{\geq}{\epsilon_{\mathrm U}}\revise{\}}.
    \end{aligned}
    \label{eq.preferenceThreshold}
\end{equation}
\revise{Here,} ${\epsilon_{\mathrm C}}$, ${\epsilon_{\mathrm B}}$, and ${\epsilon_{\mathrm U}}$ are the \revise{agent-specific} preference thresholds\revise{.} 
\revise{Among the active offers indexed by the agents in $\mathcal{F}(p)$, the control center retains the highest-scoring offer from each represented agent type under a fixed tie-breaking rule. Let $\mathcal{T}(p)$ denote the resulting set of at most three type-level candidates. If $\mathcal{T}(p)\neq\emptyset$, it provisionally selects}
\begin{equation*}
\revise{a^\star(p)\in\arg\max_{a\in\mathcal{T}(p)}
\frac{\rho(a,p)}{s(a,p)},}
\end{equation*}
\revise{This deterministic dispatch rule combines the learned preference score with marginal monetary cost through the preference-to-cost ratio and makes no global-optimality claim; residual parcels are handled separately in Sec.~\ref{sec.restParcelAssignment}. The offer is committed only after $a^\star(p)$ returns a positive acknowledgement. If the offer is rejected, it is removed and the same rule is reapplied to the remaining valid high-preference offers for $p$. If no such offer remains, $p$ is not committed in this stage.} \revise{The disjoint sets} ${\mathbb{P}_{\mathrm C}}$, ${\mathbb{P}_{\mathrm B}}$, and ${\mathbb{P}_{\mathrm U}}$ \revise{contain the positively acknowledged parcels committed} to couriers, GVs, and UAVs, respectively\revise{.} \revise{Offers are processed sequentially; after each commitment, the selected agent updates its route and status, and infeasible stale offers are removed.} \revise{The remaining set is}
$\mathbb{P}^\prime=\mathbb{P}\revise{\setminus}({\mathbb{P}_{\mathrm C}}\cup{\mathbb{P}_{\mathrm B}}\cup{\mathbb{P}_{\mathrm U}})$.
\revise{All parcels in $\mathbb{P}^{\prime}$ enter the remaining-parcel stage below.}

\subsection{Remaining Parcel Assignment}\label{sec.restParcelAssignment}
Delivery companies \revise{place fulfillment at the highest priority because successful service directly affects customer experience and reputation. We therefore adopt a lexicographic objective rather than assigning equal priority to fulfillment and cost:} the first \revise{stage maximizes feasible fulfillment, and} the second \revise{minimizes cost without reducing that fulfillment level.} 
\revise{All assignments committed in the preceding preference-based stage are held fixed, so the objectives below apply only to the residual problem over $\mathbb{P}^{\prime}$ rather than to a joint reoptimization of the original batch.}

\revise{Let $\mathbb{A}=\mathbb{U}\cup\mathbb{C}\cup\mathbb{B}$ be the disjoint union of all agents, and let $z(a,p)\in\{0,1\}$ indicate whether remaining} parcel $p\in\mathbb{P}^{\prime}$ is assigned to \revise{agent $a\in\mathbb{A}$.} \revise{The notation $s(a,p)$ denotes the corresponding UAV, courier, or GV marginal monetary cost defined in Sec.~\ref{sec.model}. Let $\mathbb{P}(a)$ denote the parcels already committed to agent $a$ before the remaining-parcel stage. Let $\mathcal{O}\subseteq\mathbb{A}\times\mathbb{P}^{\prime}$ contain the agent--parcel keys of all currently valid, locally verified feasible offers for the remaining parcels, including offers whose preference scores do not exceed the preceding thresholds; those thresholds are not reapplied in this stage. Each key $(a,p)\in\mathcal O$ refers to the unique active offer record. For an assignment $\boldsymbol{z}$, define $\mathbb{P}_{\boldsymbol{z}}(a)=\{p\in\mathbb{P}^{\prime}\mid z(a,p)=1\}$. The feasible set $\mathcal{Z}$ contains exactly the assignments satisfying}
\begin{equation}
\begin{aligned}
\revise{z(a,p)}&\in\{0,1\}, &&\revise{a\in\mathbb{A}},\ p\in\mathbb{P}^{\revise{\prime}},\\
\sum_{\revise{a\in\mathbb{A}}}\revise{z(a,p)}&\leq1, &&p\in\mathbb{P}^{\revise{\prime}},\\
\revise{z(a,p)}&\revise{=0,} &&\revise{(a,p)\notin\mathcal{O},}\\
\revise{\operatorname{Feas}_a\!\left(\mathbb{P}(a)\cup
\mathbb{P}_{\boldsymbol{z}}(a)\right)}&\revise{=1,} &&\revise{a\in\mathbb{A}.}
\end{aligned}
\label{eq.feasibleAssignmentSet}
\end{equation}
\revise{Here, $\operatorname{Feas}_a(\mathbb P(a)\cup\mathbb P_{\boldsymbol z}(a))=1$ requires the parcels already committed to agent $a$ and all residual parcels newly assigned to it under $\boldsymbol z$ to jointly satisfy the model-specific delivery constraints.}
\revise{This set-level check prevents individually feasible offers from forming an infeasible combined route.}


\revise{For $\boldsymbol{z}\in\mathcal{Z}$, let $S_a(\mathbb{Q}\mid\mathbb{P}(a))$ denote the minimum incremental monetary cost of jointly inserting parcel set $\mathbb{Q}$ into agent $a$'s current plan while satisfying $\operatorname{Feas}_a$. We set $S_a(\emptyset\mid\mathbb{P}(a))=0$. For a single offer, $S_a(\{p\}\mid\mathbb{P}(a))=s(a,p)$. This route-level definition accounts for interactions among multiple insertions; in the implementation, its incremental counterpart is recomputed after every committed insertion. Define the assigned-parcel count and total monetary cost as}
\begin{equation}
\begin{aligned}
\revise{N(\boldsymbol{z})}&=\sum_{\revise{p\in\mathbb{P}^{\prime}}}
\sum_{\revise{a\in\mathbb{A}}}\revise{z(a,p)},\\
\revise{C(\boldsymbol{z})}&\revise{=\sum_{a\in\mathbb{A}}
S_a\!\left(\mathbb{P}_{\boldsymbol{z}}(a)\mid\mathbb{P}(a)\right).}
\end{aligned}
\label{eq.assignmentMetrics}
\end{equation}
\revise{The ideal two-stage lexicographic model is}
\begin{equation}
\revise{\begin{aligned}
N_{\max}
&=\max_{\boldsymbol{z}\in\mathcal{Z}}N(\boldsymbol{z}),
&&\text{\rm (Stage 1)}\\[2pt]
\min_{\boldsymbol{z}\in\mathcal{Z}}\quad
&C(\boldsymbol{z}),
&&\text{\rm (Stage 2)}\\
\text{s.t.}\quad
&N(\boldsymbol{z})=N_{\max}.
\end{aligned}}
\label{eq.lexicographicAssignment}
\end{equation}
\revise{With the earlier preference-based commitments fixed, any exact optimizer $\boldsymbol{z}^{\star}$ of Eq.~\eqref{eq.lexicographicAssignment} is Pareto-optimal with respect to maximizing feasible fulfillment and minimizing monetary cost within the residual feasible set $\mathcal{Z}$. Indeed, no feasible assignment can assign more than $N_{\max}$ parcels, and among the assignments attaining $N_{\max}$, none can have a lower cost than $C(\boldsymbol{z}^{\star})$. This guarantee concerns only the ideal formulation.
Because $\mathbb{A}=\mathbb{U}\cup\mathbb{C}\cup\mathbb{B}$, $C(\boldsymbol{z})$ includes the monetary delivery costs of UAVs, couriers, and GVs. An ordinary residual parcel with no feasible offer may remain unassigned in the current allocation cycle; recovery tasks instead follow the pending-and-reallocation procedure in Sec.~\ref{sec.realtimeFallback}.}

\revise{The control center updates $\mathcal{O}$ whenever an agent becomes unavailable or an offer becomes stale. If an entire class is unavailable, for example, all UAVs are grounded by severe weather, all offers from that class are removed. If only individual agents are unavailable, only their offers are removed. Each selected assignment remains provisional until the corresponding agent returns a lightweight acknowledgement that its offer remains current. For a GV, whose delivery feasibility additionally depends on its private and potentially changing passenger route, the acknowledgement includes a local consistency check of route overlap, detour tolerance, and timing constraints. A rejected offer $(a,p)$ is removed from $\mathcal{O}$, after which the residual-assignment procedure is rerun over the updated offer set for the unserved parcels. 
Following Sec.~\ref{sec.realtimeFallback}, a revoked unpicked parcel, as well as a recovery task created for a picked but undelivered parcel, is inserted into the current $\mathbb{P}^{\prime}$. For a recovery task, stale offers are invalidated and $\mathcal{O}$ is rebuilt from fresh locally verified offers using the updated pickup location and delivery-time reference specified in Sec.~\ref{sec.realtimeFallback}; all unaffected commitments remain fixed.}

\revise{For scalable city-wide implementation, we use the} greedy heuristic \revise{in~\cite{10598004,11261388}. It follows the same priority order by treating fulfillment as primary and cost as secondary, and it refreshes marginal costs and joint feasibility after each committed insertion.} 
\revise{Let $\mathcal{K}=|\mathcal{O}|\leq\mathcal{MN}$ be} the number of \revise{current offers, where $\mathcal{M}=|\mathbb{A}|$ and $\mathcal{N}=|\mathbb{P}^{\prime}|$. Excluding agent-side route-feasibility and cost-refresh calculations, sorting and scanning the offers costs $O(\mathcal{K}\log\mathcal{K})$, or $O(\mathcal{MN}\log(\mathcal{MN}))$ in the worst case~\cite{COHEN2006162}. This stage is a scalable residual-assignment method rather than a new exact solver.} The heuristic \revise{therefore approximates the ideal model in Eq.~\eqref{eq.lexicographicAssignment} and is not claimed to inherit its Pareto-optimality guarantee.}

\section{Performance Evaluation}\label{sec.evaluation}

\subsection{Evaluation Settings}

\subsubsection{Evaluation Implementation} 
\revise{We use the one-month delivery-order and taxi-trajectory datasets
described in Sec.~\ref{sec.dataDescription}. Each dataset is independently
partitioned by} day: the first 23 days \revise{are} used \revise{for model training} and
\revise{fine-tuning,} and \revise{the final seven for evaluation}. \revise{Unless stated otherwise},
results are \revise{seven-day means}, \revise{and} error bars \revise{are} standard \revise{deviations across
the seven daily values}. \revise{The default layout contains} 15 \revise{UAV} and 50 courier
stations. \revise{Based on} the DJI Mavic 3 Pro \revise{specifications}~\cite{DJIMavic3Pro},
\revise{UAV} speed \revise{and endurance are} set to $16\ \mathrm{m/s}$ \revise{(about $75\%$} of
the \revise{reported} maximum\revise{)} and 40 \revise{min (below the reported 43-min maximum),
respectively}.

\subsubsection{Evaluated Parameters}
We \revise{evaluate the effects of} four parameters \revise{that substantially influence} delivery performance and \revise{cost}.
\begin{itemize}
    \item \textbf{Delivery \revise{Demand}.} \revise{We test demand} ratios of $\{70\%,80\%,90\%,\underline{100\%}\}$ \revise{to represent} different delivery burdens.
    \item \textbf{Ratio of Taxis Participating in Delivery.} \revise{Because} not \revise{every taxi is} willing to deliver parcels, \revise{we test participating-taxi ratios} of $\{0.05,\underline{0.10},0.15,0.20\}$. 
    \item \textbf{Number of UAVs per Station.} UAV \revise{fleet size directly affects available aerial-delivery capacity}. We \revise{therefore test} $\{15,20,\underline{25},30\}$ UAVs per station.
    \item \textbf{Number of Couriers per Station.} \revise{Courier fleet size likewise affects ground-delivery capacity}. We \revise{test} $\{10,15,\underline{20},25\}$ couriers per station.
\end{itemize}
The underlined values are the default \revise{settings}.

\subsubsection{Baselines}
We compare {\sf TriDeliver} (abbreviated as T-D) against three baselines as follows.

\begin{itemize}
    \item \textbf{On-Demand Delivery (O-D).} O-D assigns \revise{each new parcel} to \revise{the feasible agent with the earliest estimated pickup time}.

    \item \textbf{Cooperative Delivery by UAVs and Crowdsourced Taxis (U-T).} U-T \revise{is} the \revise{UAV--taxi} cooperative delivery \revise{method proposed} in~\cite{10598004} \revise{and does not employ couriers}.
    
    \item \textbf{Ground Truth by Couriers (G-T)~\cite{9524841}.} G-T \revise{represents} the \revise{observed courier-only} delivery \revise{process recorded} in the \textit{aBeacon} dataset.
    
\end{itemize}

\subsubsection{Metrics}
\revise{We} evaluate T-D \revise{using the following five metrics covering} delivery \revise{efficiency, operational} cost, \revise{and agent-level economic impacts}.

\begin{itemize}
    \item \textbf{Number of Parcels \revise{Delivered}.}
    \revise{This metric measures} the \revise{total} number of parcels \revise{successfully delivered within} the \revise{specified deadline. A larger value indicates higher} delivery \revise{throughput and better} delivery performance.
    \item \textbf{Delivery Time.}
    \revise{This metric measures the average} time \revise{required to successfully deliver a parcel, with the} delivery \revise{deadline set to} 60 minutes. \revise{A smaller value indicates} faster delivery \revise{and} better \revise{service timeliness}.
\item \textbf{Delivery Cost.}
\revise{This metric is the total modeled monetary} cost \revise{of completed UAV, courier,}
and \revise{taxi assignments}. \revise{At comparable fulfillment levels}, a \revise{smaller value
indicates lower operating} cost.

\item \textbf{\revise{Taxi} Delivery Price.}
\revise{This metric} is the \revise{mean modeled compensation} per \revise{taxi-delivered} parcel\revise{.
Because it largely scales with additional travel} distance, \revise{a smaller value
generally indicates less interference with the original passenger trip}.

\item \revise{\textbf{Relative Courier Compensation.}
This metric is the modeled payment per available courier divided by the
corresponding G-T value. A value above one indicates a higher average
modeled payment than under G-T.}

\end{itemize}

\begin{table}[t]
\centering
\caption{\revise{Sensitivity of Aggregate Performance to Random Station Layouts}}
\label{tab.random_station_layout_sensitivity}
\setlength{\tabcolsep}{3pt}
\renewcommand{\arraystretch}{1.15}
\begin{tabular}{lrrr}
\hline
\revise{Statistic}
& \shortstack{\revise{Delivered}\\\revise{parcels}}
& \shortstack{\revise{Total delivery}\\\revise{cost (CNY)}}
& \shortstack{\revise{Average delivery}\\\revise{time (min)}} \\
\hline
\revise{Default layout}
& \revise{21,574.43}
& \revise{94,766.15}
& \revise{27.291} \\

\revise{Random-layout mean}
& \revise{21,567.73}
& \revise{96,452.28}
& \revise{28.005} \\

\revise{Sample SD}
& \revise{146.68}
& \revise{3,418.46}
& \revise{1.531} \\

\revise{CV}
& \revise{0.68\%}
& \revise{3.54\%}
& \revise{5.47\%} \\

\revise{Minimum}
& \revise{21,257.43}
& \revise{91,214.34}
& \revise{25.560} \\

\revise{Maximum}
& \revise{21,797.71}
& \revise{101,214.74}
& \revise{30.492} \\

\revise{Mean vs.\ default}
& \revise{$-0.03\%$}
& \revise{$+1.78\%$}
& \revise{$+2.62\%$} \\
\hline
\end{tabular}
\parbox{\columnwidth}{\footnotesize
\revise{\emph{Note:} Statistics are computed across nine layout-specific
seven-day means (15 UAV and 50 courier stations per layout), excluding
the default layout. SD is the sample standard deviation, CV is SD/mean,
and ``Mean vs.\ default'' is the percentage difference from the default.}
}
\end{table}

\begin{figure*}[t!]
\setlength{\abovedisplayskip}{2pt}
\setlength{\belowdisplayskip}{2pt}
\centering  
\subfigure[\revise{Delivered Parcels}]{
\label{fig.deliProp.1}
\includegraphics[width=0.23\textwidth]{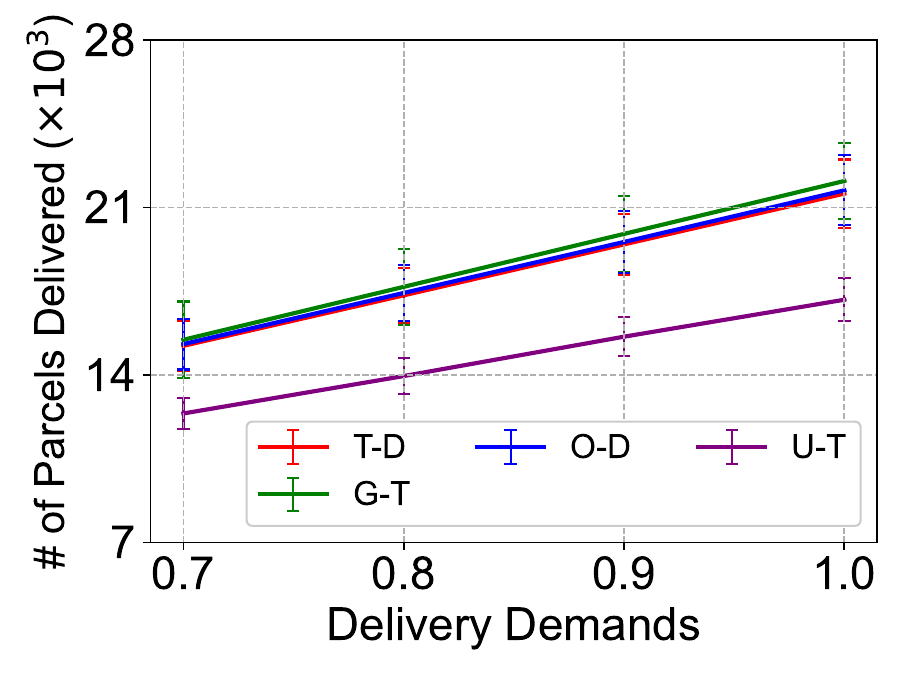}}
\subfigure[Delivery Time]{
\label{fig.deliProp.2}
\includegraphics[width=0.23\textwidth]{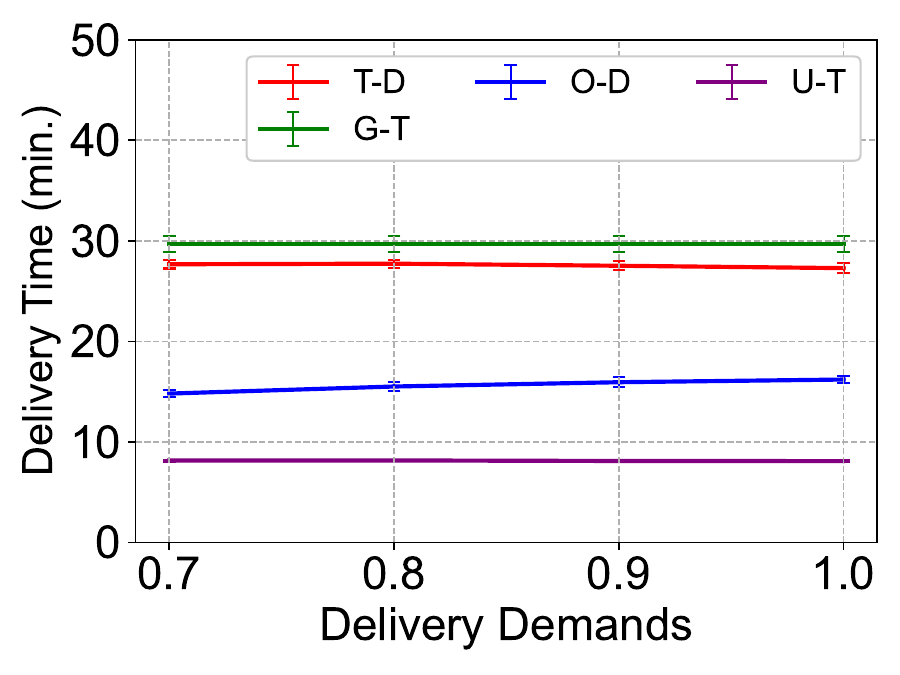}}
\subfigure[Delivery Cost]{
\label{fig.deliProp.3}
\includegraphics[width=0.23\textwidth]{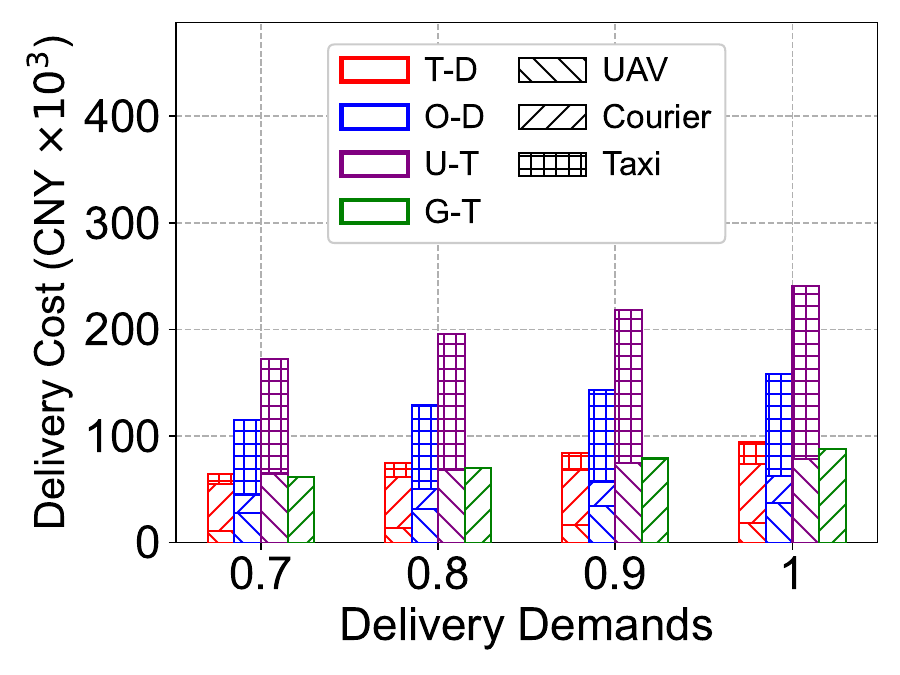}}
\subfigure[\revise{Taxi} Price \revise{and Courier Compensation}]{
\label{fig.deliProp.4}
\includegraphics[width=0.25\textwidth]{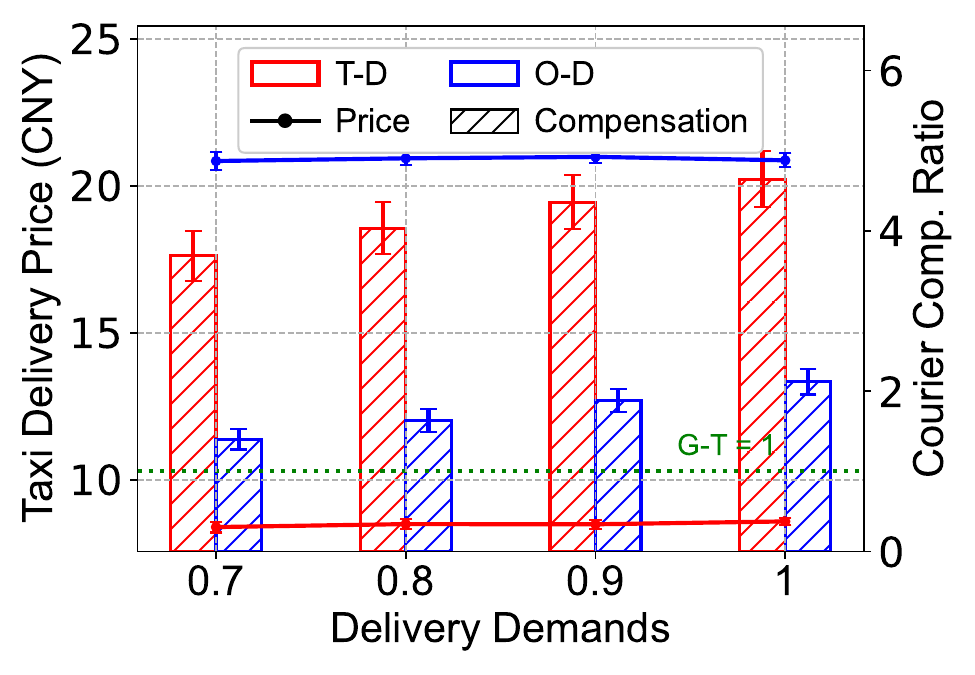}}
\vspace{-0.2cm}
\caption{Impact of Delivery Demands on Delivery Performance}
\label{fig.deliProp}
\vspace{-0.3cm}
\end{figure*}

\begin{figure*}[t!]
\setlength{\abovedisplayskip}{2pt}
\setlength{\belowdisplayskip}{2pt}
    \centering  
\subfigure[\revise{Delivered Parcels}]{
    \label{fig.taxiProp.1}
    \includegraphics[width=0.23\textwidth]{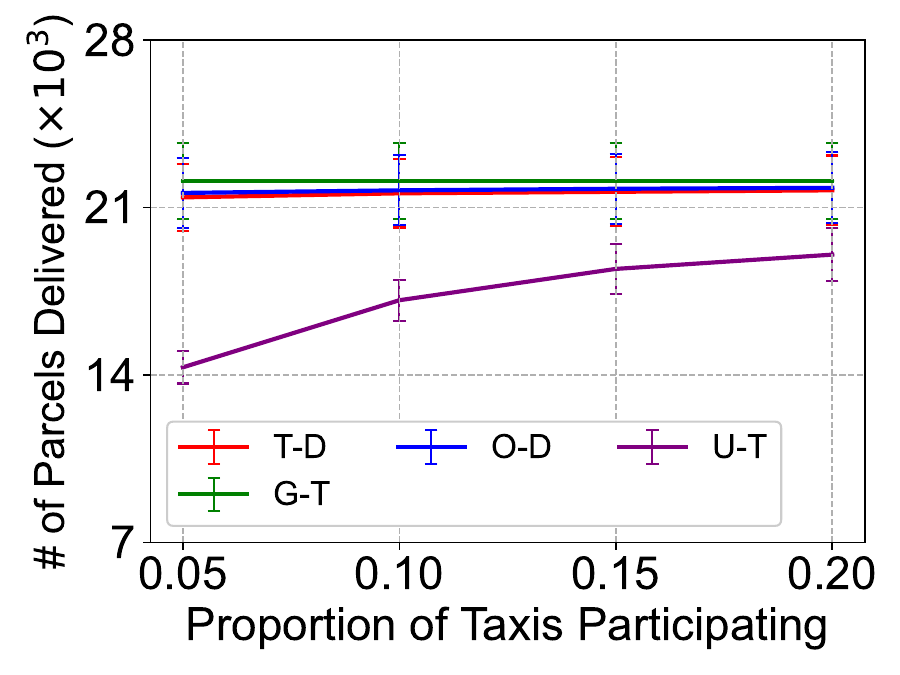}}
\subfigure[Delivery Time]{
    \label{fig.taxiProp.2}
    \includegraphics[width=0.23\textwidth]{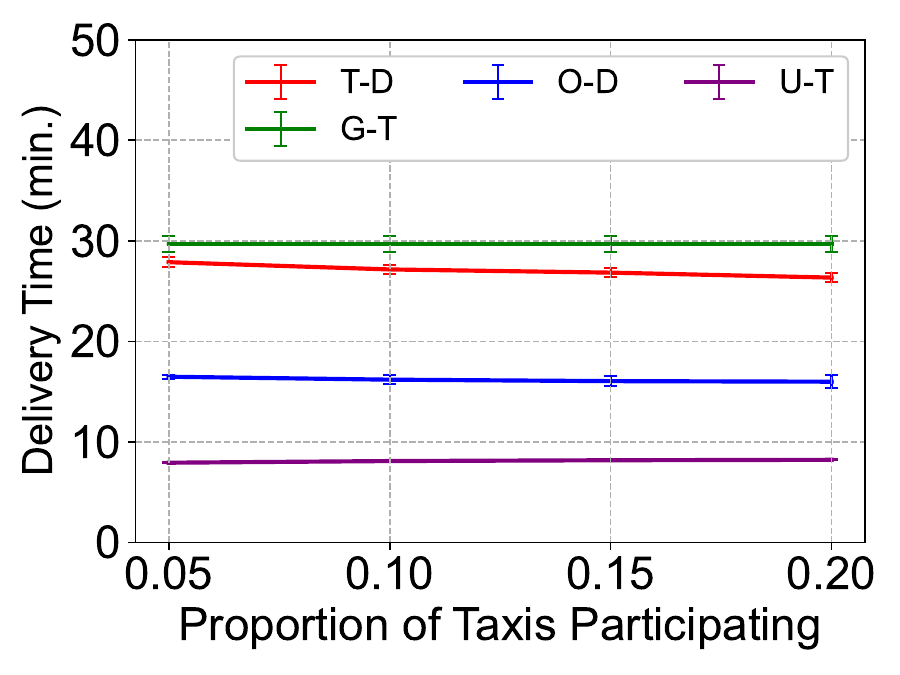}}
\subfigure[Delivery Cost]{
    \label{fig.taxiProp.3}
    \includegraphics[width=0.23\textwidth]{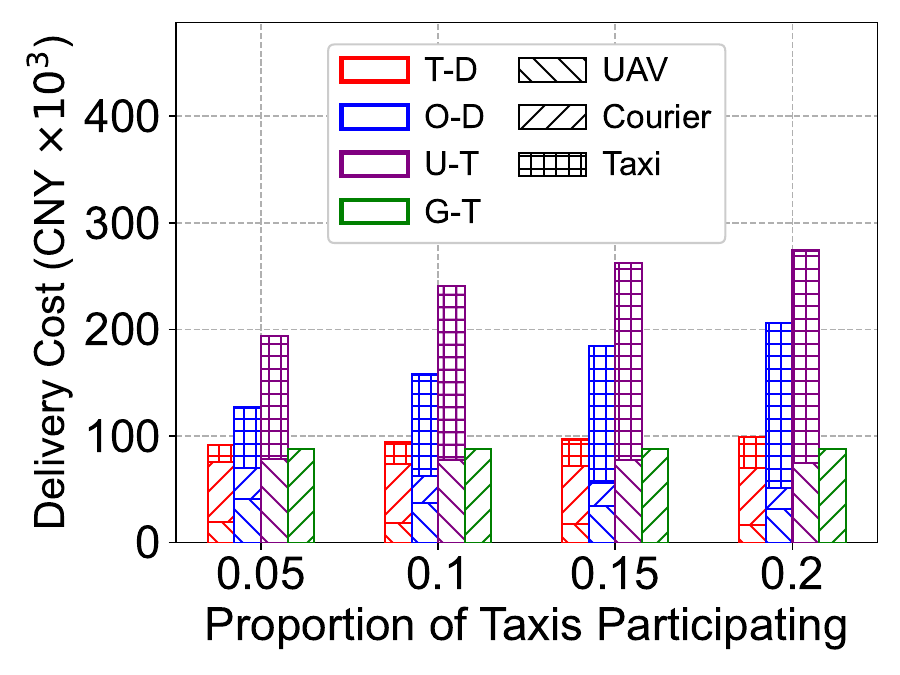}}
\subfigure[\revise{Taxi} Price \revise{and Courier Compensation}]{
    \label{fig.taxiProp.4}
    \includegraphics[width=0.26\textwidth]{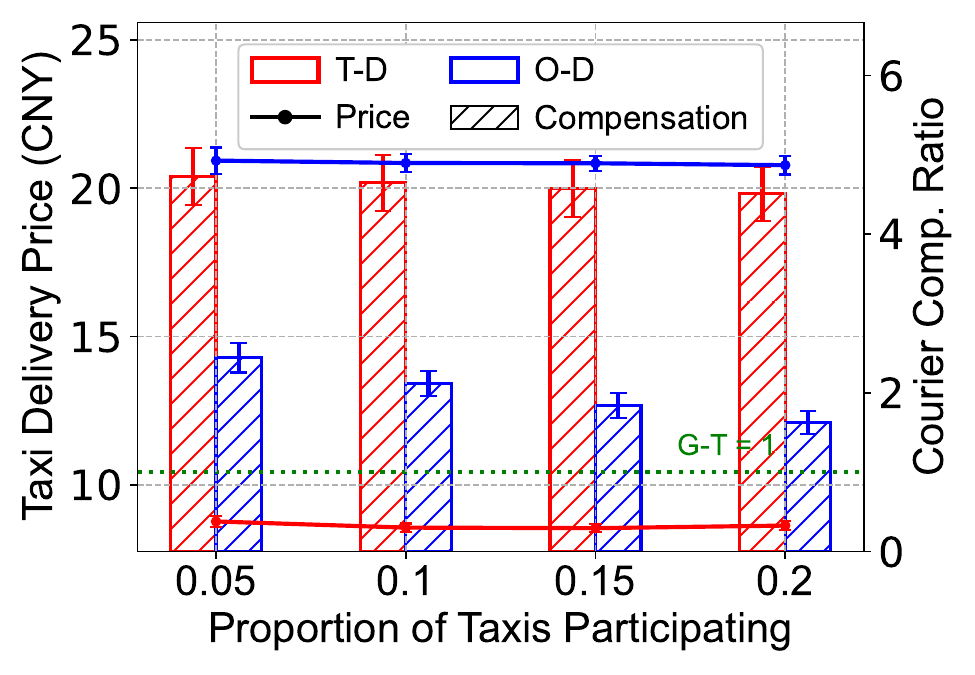}}
    \vspace{-0.2cm}
    \caption{Impact of Ratio of Taxis Participating in Delivery on Delivery Performance}
    \label{fig.taxiProp}
    \vspace{-0.3cm}
\end{figure*}

\begin{figure*}[t]
\setlength{\abovedisplayskip}{2pt}
\setlength{\belowdisplayskip}{2pt}
    \centering  
    \subfigure[\revise{Delivered Parcels}]{
    \label{fig.UAVNum.1}
    \includegraphics[width=0.23\textwidth]{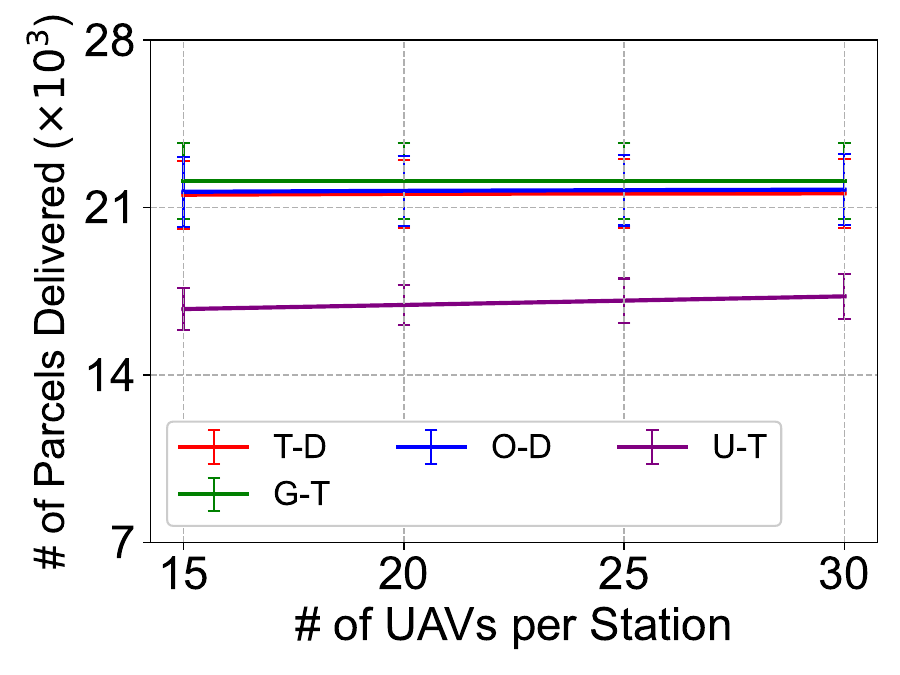}}
    \subfigure[Delivery Time]{
    \label{fig.UAVNum.2}
    \includegraphics[width=0.23\textwidth]{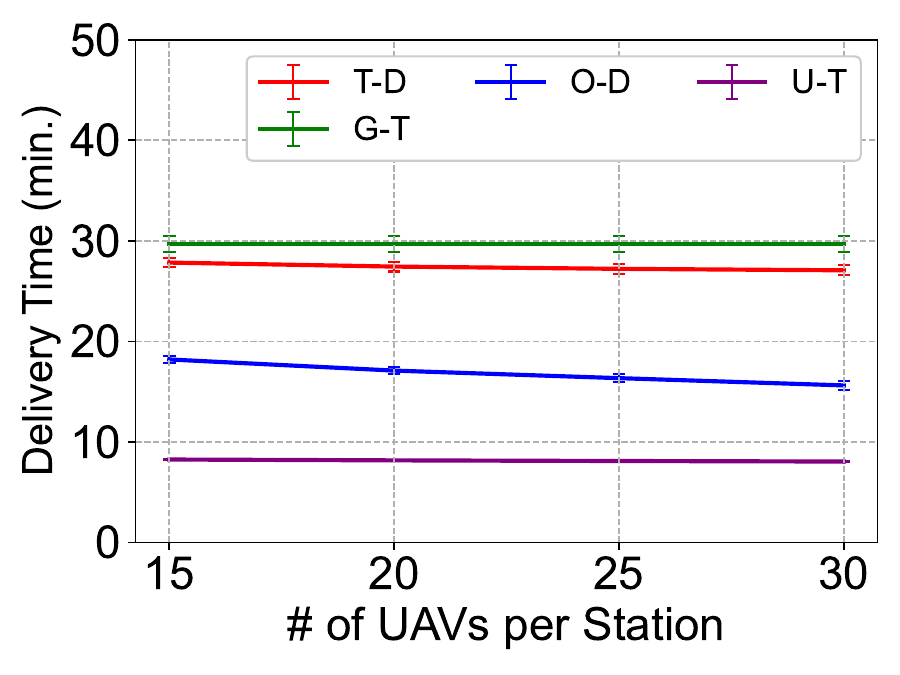}}
    \subfigure[Delivery Cost]{
    \label{fig.UAVNum.3}
    \includegraphics[width=0.23\textwidth]{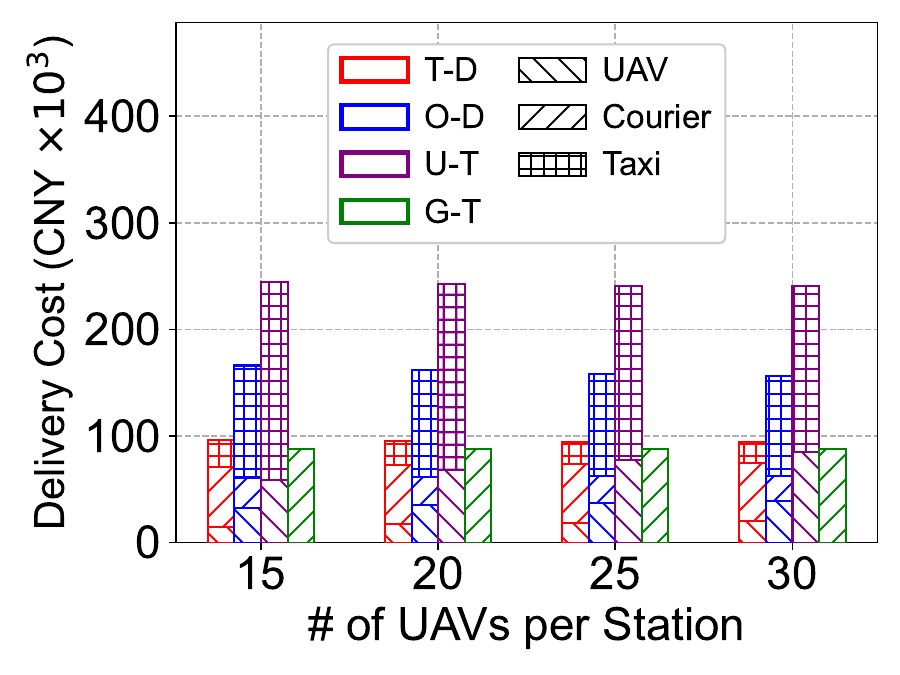}}
    \subfigure[\revise{Taxi} Price \revise{and Courier Compensation}]{
    \label{fig.UAVNum.4}
    \includegraphics[width=0.26\textwidth]{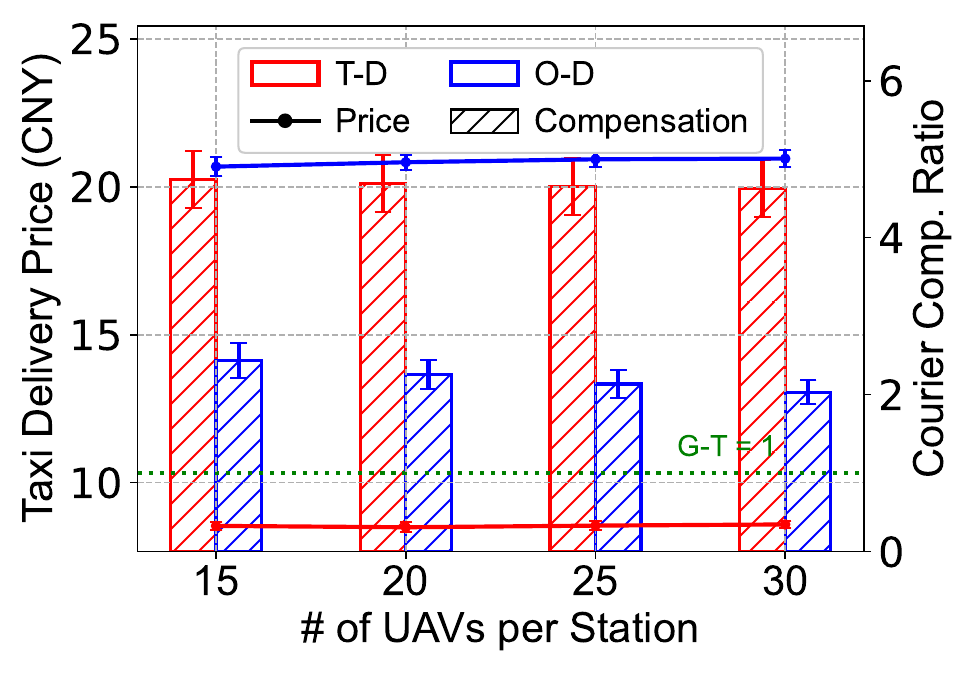}}
    \vspace{-0.2cm}
    \caption{Impact of the Number of UAVs at Each Station on Delivery Performance}
    \label{fig.UAVNum}
    \vspace{-0.3cm}
\end{figure*}

\begin{figure*}[t!]
\setlength{\abovedisplayskip}{2pt}
\setlength{\belowdisplayskip}{2pt}
    \centering  
    \subfigure[\revise{Delivered Parcels}]{
    \label{fig.courierNum.1}
    \includegraphics[width=0.23\textwidth]{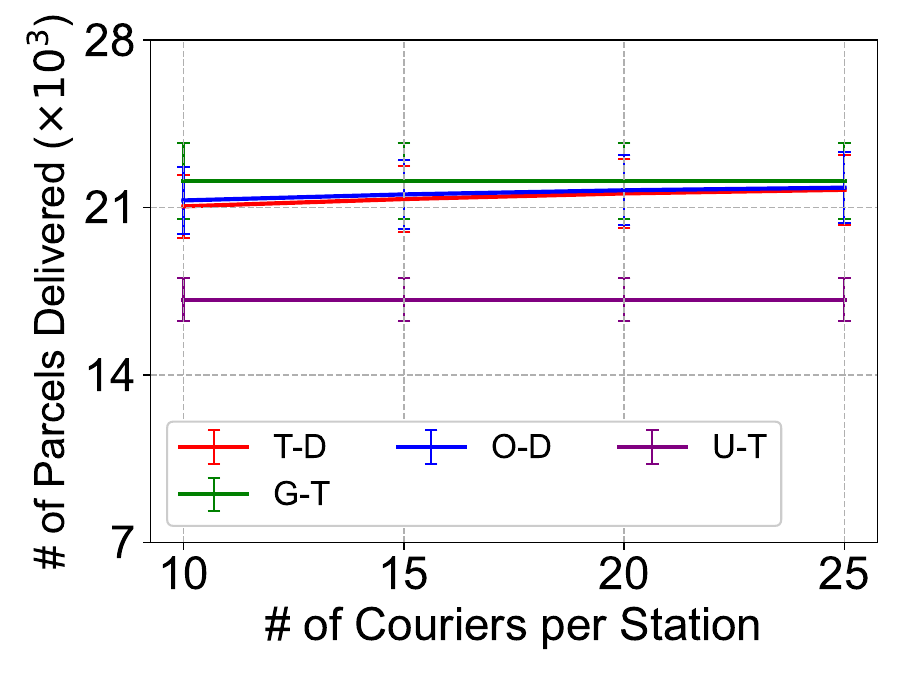}}
    \subfigure[Delivery Time]{
    \label{fig.courierNum.2}
    \includegraphics[width=0.23\textwidth]{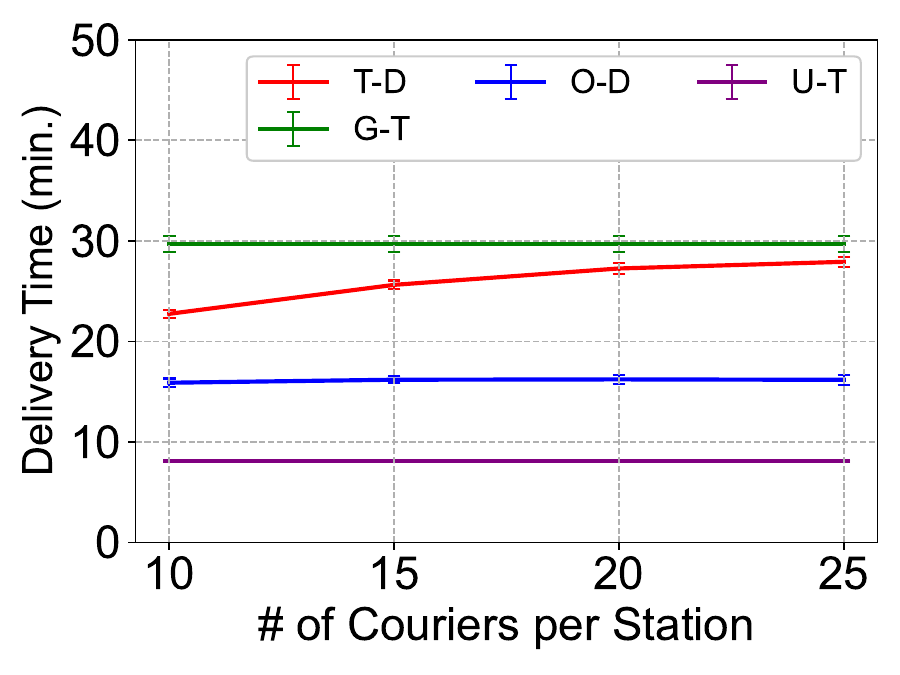}}
    \subfigure[Delivery Cost]{
    \label{fig.courierNum.3}
    \includegraphics[width=0.23\textwidth]{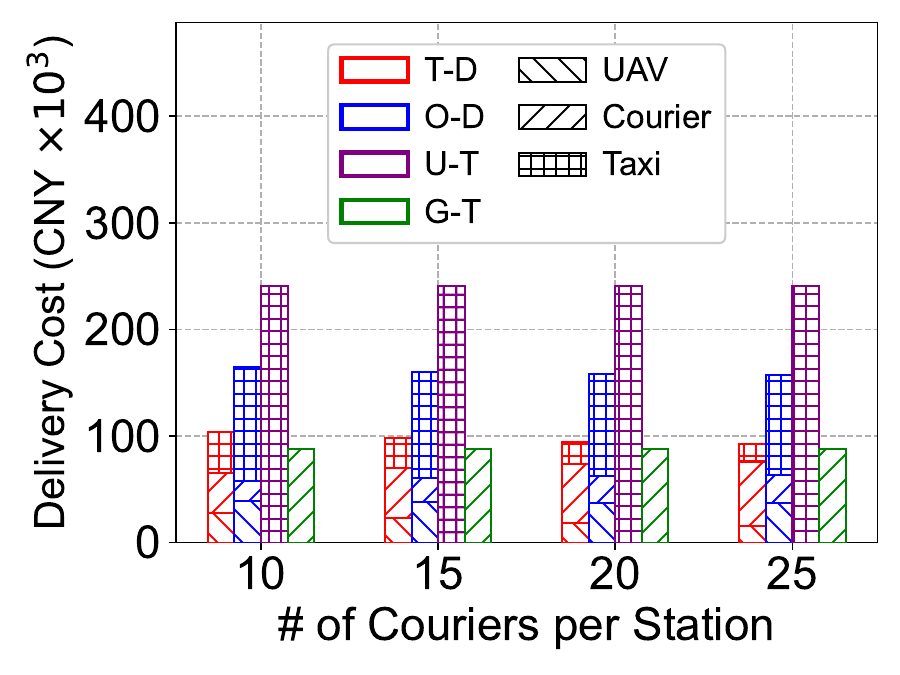}}
    \subfigure[\revise{Taxi} Price \revise{and Courier Compensation}]{
    \label{fig.courierNum.4}
    \includegraphics[width=0.26\textwidth]{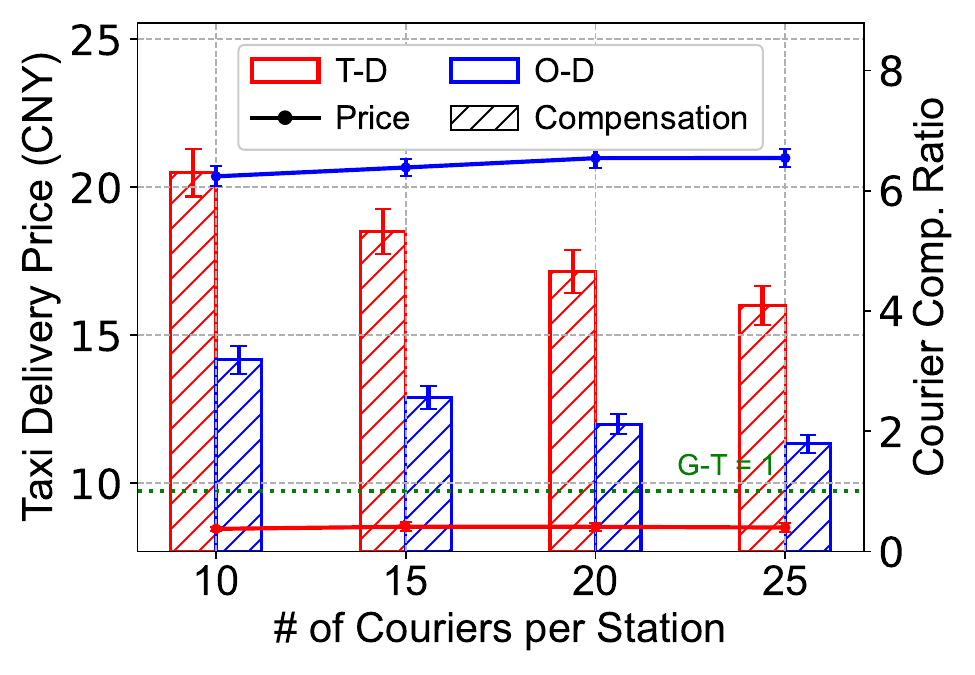}}
    \vspace{-0.2cm}
    \caption{Impact of the Number of Couriers at Each Station on Delivery Performance}
    \label{fig.courierNum}
    \vspace{-0.3cm}
\end{figure*}

\subsection{\revise{Sensitivity to Random Station Layouts}}
\revise{To evaluate the sensitivity of {\sf TriDeliver} to randomly generated UAV and courier station locations, we generated nine additional station layouts using nine independent random seeds and the same station-generation procedure as the default layout. Each layout contains 15 UAV stations and 50 courier stations and was evaluated over the same seven days using the default parameter settings. The results are shown in Table~\ref{tab.random_station_layout_sensitivity}.
Across the nine random layouts, aggregate performance remained stable: the CVs of delivered parcels, total delivery cost, and average delivery time were $0.68\%$, $3.54\%$, and $5.47\%$, respectively, while their means differed from the default layout by only $-0.03\%$, $+1.78\%$, and $+2.62\%$. For all three aggregate metrics, the default-layout result was within 0.5 sample SD of the corresponding random-layout mean. Mode allocation was more sensitive, with parcel-count CVs of $31.88\%$, $6.74\%$, and $18.37\%$ for UAVs, couriers, and taxis, respectively. Thus, TriDeliver primarily adapts to station placement by reallocating work among agent types while maintaining comparatively stable aggregate performance.}



\subsection{Ablation Study}

\begin{table}[t!]
    \centering
    \caption{Results of Ablation Study.}
    \scriptsize
    \setlength{\tabcolsep}{2.2pt}
\begin{tabular}{lrrrr}
\hline
 &\makecell{\revise{Delivered}\\\revise{parcels}}&\makecell{\revise{Avg.} delivery\\time \revise{(min)}}
 &\makecell{\revise{Total} delivery\\cost \revise{(CNY)}}&\makecell{\revise{Taxi} delivery price\\\revise{(CNY/parcel)}}\\
\hline
T-D     &\revise{21,574} &\revise{27.29} &\revise{94,766}  &\revise{8.583}\\
w/o TL  &\revise{21,601} &\revise{28.94} &\revise{100,637} &\revise{10.853}\\
\hline
\end{tabular}
\label{tab.ablationStudy}
\end{table}

To \revise{isolate} the \revise{contribution of} transfer learning, we compare \revise{T-D} with \revise{an} ablation variant \revise{without TL} (w/o TL). \revise{As} shown in Table~\ref{tab.ablationStudy}\revise{, both variants deliver approximately $97.6\%$ of the orders}. \revise{Relative} to w/o TL\revise{, T-D delivers only $0.1\%$ fewer parcels but reduces average delivery time, total modeled cost, and taxi delivery price by $5.7\%$, $5.8\%$, and $20.9\%$, respectively}. \revise{These results indicate that the transferred and target-specific networks improve offer ranking. In particular, the lower taxi price is consistent with avoiding high-detour GV assignments.}



\subsection{Numerical Results}
\subsubsection{Impacts of Delivery Demands}
Fig.~\ref{fig.deliProp} illustrates the impacts of delivery demands on the \revise{five evaluation metrics}. As shown in Fig.~\ref{fig.deliProp.1}, \revise{the numbers of} parcels delivered \revise{by all} methods increase \revise{with delivery demands}. T-D \revise{consistently} delivers more than \revise{$97.6\%$ of all parcels ordered}. \revise{At $100\%$ demand}, T-D delivers \revise{21,574 parcels}, which is \revise{only $0.6\%$ fewer than} O-D \revise{but $25.9\%$} more than U-T. This \revise{result demonstrates that the heterogeneous delivery resources coordinated by} T-D \revise{provide} sufficient capacity \revise{under heavy demand}. \revise{In contrast, without courier participation, the capacity of U-T is constrained by the available UAVs and crowdsourced taxis.}
Fig.~\ref{fig.deliProp.2} \revise{shows that the delivery time of} T-D \revise{remains stable between 27.3 and 27.7 minutes as the demand increases}. \revise{Its average} delivery time is \revise{27.56} minutes, which \revise{is $7.2\%$} shorter than that \revise{of} G-T. O-D \revise{and U-T achieve shorter} average delivery \revise{times} of \revise{15.62 and 8.13} minutes\revise{, respectively}. \revise{O-D explicitly prioritizes the earliest pick-up, whereas U-T relies only on fast UAVs and taxis with limited detours. By comparison, T-D screens offers using feasibility constraints and learned preferences, and then prioritizes fulfillment and modeled monetary cost; it is therefore not designed to minimize delivery time alone.}
\revise{As shown in Fig.~\ref{fig.deliProp.3}}, \revise{a} heavier delivery burden \revise{increases the total} delivery \revise{cost of every method}. When \revise{the} demand \revise{grows from $70\%$ to $100\%$, the cost of} T-D \revise{rises by $47.1\%$, from 64,413 CNY to 94,766 CNY, because more parcels must be assigned across UAVs, couriers, and taxis}. \revise{Nevertheless, at} $100\%$ \revise{demand}, T-D \revise{reduces} the delivery cost by \revise{$40.0\%$} and \revise{$60.7\%$ compared with} O-D \revise{and U-T, respectively}. \revise{Although G-T has the lowest total cost, the cost of T-D is only $5.0\%$--$8.1\%$ higher across the four demand levels while involving substantially fewer couriers.
Fig.~\ref{fig.deliProp.4} further reports the taxi delivery price using curves and the left axis, and the courier compensation ratio relative to G-T using bars and the right axis. The average taxi delivery price of T-D remains stable between 8.39 and 8.58 CNY per parcel across all demand levels. Its mean} price is \revise{8.49} CNY, which is \revise{$59.4\%$} and \revise{$49.9\%$} lower than \revise{those of} O-D and U-T, respectively. \revise{This indicates that} the \revise{preference-aware coordination} in T-D \revise{avoids assigning} parcels to taxis \revise{requiring long} detours\revise{, thereby limiting the impact on their passengers}. \revise{Meanwhile, the courier compensation ratio of T-D increases from 3.69 to 4.65 as demand grows, whereas that of O-D increases from only 1.40 to 2.12, with G-T normalized to 1. Therefore, T-D provides couriers with substantially higher compensation while maintaining a low taxi delivery price and a competitive overall delivery cost.}

\subsubsection{Impacts \revise{of} Ratio of Taxis Participating in Delivery}
Fig.~\ref{fig.taxiProp.1} \revise{shows that increasing} the \revise{ratio of participating taxis from $5\%$ to $20\%$ raises} the number of parcels delivered by \revise{T-D only slightly, from 21,417 to 21,712} (\revise{$+1.4\%$}). \revise{A similarly small increase of $1.0\%$ is observed} for \revise{O-D because both methods already have sufficient heterogeneous} delivery \revise{capacity at} the \revise{lowest} taxi ratio. \revise{By contrast}, the number of parcels delivered \revise{by} U-T increases \revise{from 14,320 to 19,026} (\revise{$+32.9\%$}), \revise{as its capacity depends directly on the available UAVs and} taxis \revise{without support from couriers}. \revise{Even at the $20\%$ taxi ratio}, T-D delivers \revise{$14.1\%$ more parcels} than U-T \revise{and reaches $98.2\%$} of the number \revise{delivered by G-T}. \revise{These results indicate that T-D benefits from additional taxis but does not rely heavily on them to maintain a high fulfillment level.
As shown in} Fig.~\ref{fig.taxiProp.2}, \revise{a larger candidate taxi pool reduces} the \revise{average} delivery time \revise{of} T-D by \revise{$5.5\%$, from 27.88 to 26.35 minutes}. \revise{More participating taxis provide spatially distributed intention offers, allowing the control center to select taxis whose original trips are better aligned with parcel routes. Thus, at the $20\%$ ratio, T-D is $11.2\%$ faster than G-T. O-D also reduces its} delivery time \revise{by $3.0\%$, reaching 16.00} minutes\revise{, because its earliest-pick-up rule can select from more nearby taxis}. In contrast, the \revise{delivery time} of U-T increases \revise{slightly} by \revise{$3.6\%$}, \revise{from 7.94 to 8.23 minutes}, \revise{because} the \revise{additional taxis enable} U-T \revise{to ship $32.9\%$ more} parcels, \revise{including marginal parcels with less favorable routes}. \revise{Nevertheless,} U-T \revise{remains the fastest method because it uses} only \revise{high-speed} UAVs and taxis \revise{under} strict detour \revise{constraints}.
Fig.~\ref{fig.taxiProp.3} \revise{illustrates how} the \revise{additional taxi capacity changes both} the \revise{composition and the total} delivery cost. \revise{In T-D, when} the \revise{taxi ratio increases from $5\%$ to $20\%$,} the \revise{taxi} delivery cost \revise{rises by $73.1\%$, while the UAV and courier costs decrease by $13.9\%$ and $4.5\%$, respectively}. \revise{This workload rebalancing limits the increase in the total cost of T-D to $7.5\%$, from 91,977 CNY to 98,919 CNY.} In \revise{comparison}, the \revise{total costs of} O-D and U-T \revise{increase by $62.9\%$ and $41.6\%$}, respectively\revise{, because both methods exploit much of the additional taxi capacity:} O-D \revise{favors the} densely distributed \revise{taxis for early} pick-up, \revise{whereas} the \revise{increased fulfillment} of U-T \revise{requires substantially more taxi deliveries}. \revise{Consequently}, \revise{at} the \revise{$20\%$ ratio,} T-D \revise{costs $52.1\%$} and \revise{$63.9\%$ less than} O-D and U-T, respectively\revise{, and only $12.8\%$ more than G-T}.
\revise{According to} Fig.~\ref{fig.taxiProp.4}, the \revise{taxi delivery} price \revise{of T-D remains stable between 8.54 and 8.76} CNY per parcel \revise{as the taxi ratio changes}. \revise{At} the \revise{$20\%$ ratio, its} price is \revise{8.62 CNY, which is $58.5\%$} and \revise{$48.7\%$ lower than} those \revise{of} O-D and U-T, respectively. \revise{This is because the fine-tuned GV preference function accounts for the original-trip status, detour burden, and marginal delivery cost, enabling T-D to exploit the larger taxi pool without selecting taxis that require long detours. Meanwhile, as some workload shifts from couriers to taxis, the courier compensation ratio of T-D decreases modestly from 4.72 to 4.51 ($-4.5\%$). The corresponding ratio of O-D drops more sharply from 2.44 to 1.62 ($-33.6\%$). Therefore, at the $20\%$ taxi ratio, T-D still provides couriers with compensation that is 4.51 times that of G-T and 2.78 times that of O-D, while keeping the impact on taxi passengers low.}

\subsubsection{Impacts \revise{of} the Number of UAVs per Station}
As shown in Fig.~\ref{fig.UAVNum.1}, \revise{increasing} the number of UAVs \revise{per} station from \revise{15} to \revise{30 changes} the numbers of parcels \revise{shipped} by T-D \revise{and} O-D \revise{by only $+0.3\%$} and \revise{$+0.4\%$}, respectively. \revise{Both methods already have sufficient heterogeneous delivery capacity with 15 UAVs per station, while adding UAVs at the existing stations increases local capacity but does not expand their geographical coverage or remove no-fly-zone constraints. U-T benefits more because it has no courier support, but its shipped volume still grows by only $3.2\%$.} With \revise{30} UAVs \revise{per} station, T-D \revise{ships 21,583 parcels}, reaching \revise{$97.7\%$} of \revise{the number shipped by} G-T. This is \revise{only $0.7\%$ fewer} than O-D \revise{but $24.9\%$ more than} U-T.
Fig.~\ref{fig.UAVNum.2} \revise{shows that additional} UAVs \revise{reduce} the \revise{average} delivery \revise{times of all three methods involving UAV delivery}. The delivery \revise{time} of T-D \revise{decreases} by \revise{$2.8\%$}, \revise{from 27.84 to 27.07 minutes, and is $8.8\%$ shorter than that of G-T when} each station \revise{has 30 UAVs}. \revise{The improvement remains moderate because additional UAV capacity can serve only agent--parcel pairs that pass the legal-path, payload, endurance, and delivery-time feasibility checks; preference scores are evaluated only for feasible pairs. O-D achieves a larger reduction of $14.2\%$, reaching 15.63 minutes, because its earliest-pick-up rule directly benefits from having more high-speed UAVs available near the stations.} The delivery time \revise{of} U-T \revise{decreases} by \revise{$2.4\%$} to \revise{8.06 minutes}. \revise{Its improvement is partly offset by} the \revise{additional} parcels \revise{enabled by the enlarged UAV fleet}, \revise{but it remains the fastest method because it uses only} UAVs and taxis.
Fig.~\ref{fig.UAVNum.3} \revise{further demonstrates that} the \revise{larger} UAV \revise{fleet redistributes} the delivery \revise{workload}. \revise{In} T-D\revise{,} the \revise{UAV} delivery cost \revise{increases} by \revise{$32.2\%$ as more parcels are served by} UAVs\revise{, whereas the courier and taxi delivery costs decrease by $2.4\%$ and $19.7\%$, respectively}. 
\revise{With UAV lifecycle operating cost included, shifting workload from taxis
to UAVs reduces T-D's total modeled cost by $1.5\%$, from 96,059 CNY to 94,595 CNY. The total costs} of O-D \revise{and} U-T \revise{decrease by $6.0\%$} and \revise{$1.6\%$}, respectively. At \revise{30 UAVs per} station, T-D \revise{costs $39.4\%$} and \revise{$60.7\%$ less} than O-D and U-T, respectively\revise{, while remaining only $7.9\%$ more expensive than} G-T. \revise{These results reflect the preference-based assignment of T-D, which retains the highest-scoring feasible offer from each agent type and then balances preference and marginal monetary cost through the preference-to-cost ratio when multiple types compete for the same parcel.
Fig.~\ref{fig.UAVNum.4} reports the taxi delivery price using curves and the left axis, and the courier compensation ratio relative to G-T using bars and the right axis. Although more UAVs reduce} the \revise{overall workload assigned to} taxis, the \revise{taxi} delivery price \revise{of} T-D \revise{remains within a narrow range of 8.49--8.58} CNY per parcel. \revise{This is because} the number of UAVs \revise{changes the amount of taxi delivery but does not materially change the alignment between taxi passenger trips} and \revise{the} parcels assigned to them. \revise{At 30 UAVs per station,} the \revise{taxi} delivery \revise{price} of T-D \revise{is 8.58} CNY, \revise{which} is \revise{$59.1\%$} and \revise{$49.4\%$} lower than those of O-D and U-T, respectively. \revise{Consistent with the workload shift toward UAVs, the courier compensation ratio of T-D decreases slightly from 4.74 to 4.63 ($-2.4\%$), while that of O-D drops from 2.44 to 2.03 ($-16.7\%$). T-D therefore still compensates couriers 4.63 times as much as G-T and 2.28 times as much as O-D at the largest UAV fleet size.}

\subsubsection{Impacts \revise{of} the Number of Couriers per Station}
As \revise{shown} in Fig.~\ref{fig.courierNum.1}, increasing \revise{the} number of couriers \revise{per station from 10 to 25 raises} the number of parcels shipped \revise{by} T-D \revise{from 21,042 to 21,733 ($+3.3\%$), while that of} O-D \revise{increases by $2.5\%$}. \revise{A larger courier pool generates more feasible intention offers and alleviates the local capacity constraints of the existing stations, enabling both methods to serve parcels that would otherwise remain unassigned. With 25} couriers \revise{per} station (1,250 couriers in total), \revise{T-D reaches $98.3\%$} of the number \revise{shipped by G-T, ships only $0.4\%$ fewer} parcels \revise{than O-D, and ships $26.8\%$ more than U-T}. Since U-T does not \revise{use couriers}, \revise{all} of \revise{its results remain fixed at} the \revise{default-setting} values \revise{throughout this experiment}.
Fig.~\ref{fig.courierNum.2} \revise{shows the accompanying fulfillment--time trade-off}. The delivery \revise{time of} T-D \revise{increases from 22.75 to 27.92 minutes} (\revise{$+22.7\%$}) \revise{because} the \revise{enlarged courier capacity both increases} the number of \revise{delivered parcels and shifts more assignments from faster UAVs and taxis to} couriers. \revise{Nevertheless, at 25 couriers per station, T-D remains $5.9\%$ faster than G-T.} The delivery time of O-D \revise{changes by only $+1.8\%$, from 15.89 to 16.18 minutes:} the shorter pick-up distance \revise{made possible by a larger candidate courier pool is largely offset by the additional, less favorable} parcels \revise{that} O-D \revise{can complete}. \revise{Thus, the observed time changes should be interpreted together with the increased fulfillment rather than as a comparison under an unchanged set of delivered parcels.}
As illustrated in Fig.~\revise{\ref{fig.courierNum.3}, the enlarged courier pool substantially redistributes the workload} in T-D. \revise{From 10 to 25 couriers per station}, \revise{courier} delivery cost increases by \revise{$62.4\%$, whereas UAV and} taxi delivery \revise{costs decrease by $43.8\%$ and $58.1\%$, respectively}. \revise{Because courier delivery substitutes for many comparatively expensive taxi assignments, the total cost of T-D falls by $11.2\%$, from 103,986 CNY to 92,304 CNY, even though it delivers more parcels. The total cost of O-D decreases by only $4.7\%$ over the same range. At 25 couriers per station, T-D costs $41.3\%$ and $61.7\%$ less than O-D and U-T, respectively, and only $5.3\%$ more than G-T. This cost pattern is consistent with the preference-aware assignment: agents first screen parcels according to their operational constraints, after which the control center retains the highest-scoring feasible offer from each agent type and applies the preference-to-cost ratio when multiple types compete for the same parcel.
Fig.~\ref{fig.courierNum.4} reports the taxi delivery price using curves and the left axis, and the courier compensation ratio relative to G-T using bars and the right axis. Although} the \revise{increased courier capacity sharply reduces the aggregate workload assigned to} taxis, the \revise{taxi} delivery price of T-D \revise{remains stable between 8.46 and 8.53} CNY \revise{per parcel}. This is because the number of couriers \revise{changes how many} parcels \revise{require taxi delivery but does not materially change the route alignment of the remaining taxi assignments}. \revise{At 25 couriers per station}, the taxi delivery \revise{price of} T-D \revise{is 8.51 CNY, which is $59.5\%$ and $49.8\%$ lower than those} of \revise{O-D and U-T, respectively}. \revise{Meanwhile, the courier compensation ratio of T-D decreases from 6.30 to 4.09 ($-35.0\%$), and that of O-D decreases from 3.19 to 1.79 ($-43.8\%$). Although the aggregate courier delivery cost of T-D increases, it is distributed among 500 to 1,250 couriers, causing the compensation per courier to decrease. Even at} the \revise{largest courier pool,} T-D still \revise{compensates couriers 4.09 times as much as} G-T and \revise{2.29 times as much as} O-D.

\section{Conclusion \revise{and Future Work}}
\label{sec.conclusion}

In this paper, we presented {\sf TriDeliver}, the first integrated \revise{air--ground} cooperative \revise{instant-delivery framework coordinating} couriers, UAVs, and \revise{crowdsourced} GVs \revise{through agent-side offers and control-center selection}. \revise{Its TL-based method learns parcel-assignment preferences from courier histories, adapts them to GVs and UAVs through feature remapping, fine-tuning, and a UAV-specific network, ranks locally verified offers by preference and marginal cost, and assigns residual parcels through lexicographic fulfillment--cost optimization.
A city-scale simulation combining month-long} real-world delivery \revise{and taxi} datasets \revise{shows} that\revise{, at full demand under the default configuration,} {\sf TriDeliver} \revise{fulfills $97.6\%$ of orders}, \revise{delivers $25.9\%$ more parcels than the} state-of-the-art \revise{UAV--taxi baseline (U-T}), and reduces \revise{total modeled} delivery cost by \revise{$60.7\%$ relative to U-T}. \revise{Relative to w/o TL}, TL reduces \revise{average} delivery time, \revise{total modeled} cost, and \revise{modeled} taxi \revise{delivery price by $5.7\%$}, \revise{$5.8\%$}, and \revise{$20.9\%$, respectively,} with \revise{essentially unchanged fulfillment}.



\revise{Because the simulation combines nonsynchronized Shanghai order and taxi datasets, with taxis serving as a proxy for ride-hailing-style crowdsourced GVs, the results do not directly generalize to private vehicles or non-urban deployment. Sparse GV participation or longer distances may reduce feasible GV and UAV offers, shift more work to couriers, and increase delivery time or modeled cost. The feasibility and fallback mechanisms still apply, but parcels remain unassigned when no agent is feasible. Future evaluation should use synchronized private-vehicle and order data across diverse urban and non-urban regions and examine long-term GV participation.}


\revise{Looking ahead, we will investigate three technical extensions. First, distribution-alignment and adversarial domain-adaptation methods could further reduce cross-agent feature discrepancies beyond the current lightweight, deployment-friendly combination of feature remapping, fine-tuning, and the UAV-specific network.}

\revise{Second, inspired by UGV-mounted reconfigurable intelligent surfaces for wireless power transfer and relaying~\cite{11224006}, we will investigate mobile UAV energy replenishment. UGVs could serve as mobile energy/relay platforms and rendezvous with UAVs for wireless charging or battery swaps, reducing detours to static depots and improving throughput and reliability.}

\revise{Finally, we will explore how 5G-Advanced/6G and advanced AI can strengthen {\sf TriDeliver}'s hierarchical edge--center architecture. In the current design, agents locally verify offers, the control center selects assignments, and selected agents revalidate offers before commitment; GVs additionally recheck route overlap, detour tolerance, and timing. Lower-latency, more reliable connectivity could accelerate offer updates, abnormal-state reporting, and assignment acknowledgements in dense deployments. Edge AI could improve travel-time, energy, detour, and feasibility estimates; multi-agent reinforcement learning could support adaptive cooperation under dynamic demand and traffic conditions; and LLM-based tools could help operators interpret system states and handle anomalies. These extensions complement the current TL-based framework.}

\bibliographystyle{IEEEtran}
\bibliography{9-reference}

@TechReport{CAITECInstantRetail2024,
  author      = {{Chinese Academy of International Trade and Economic Cooperation}},
  title       = {Instant Retail Industry Development Report (2024)},
  institution = {Ministry of Commerce of the People's Republic of China},
  year        = {2024},
  month       = oct,
  url={https://caitec.org.cn/upfiles/file/2024/11/20241213113011188.pdf},
  note        = {Original title in Chinese; accessed Jul. 27, 2026},
}

@TechReport{MeituanAnnualReport2023,
  author      = {{Meituan}},
  title       = {Annual Report 2023},
  institution = {Meituan},
  year        = {2024},
  url={https://www1.hkexnews.hk/listedco/listconews/sehk/2024/0429/2024042900115.pdf},
  note        = {Accessed Jul. 27, 2026},
}

@Online{MOHRSSShortage2022,
  author = {{Ministry of Human Resources and Social Security of the People's Republic of China}},
  title  = {Ranking of the 100 Occupations with the Greatest Recruitment Shortages in the Fourth Quarter of 2022},
  year   = {2023},
  url={https://www.mohrss.gov.cn/SYrlzyhshbzb/dongtaixinwen/buneiyaowen/rsxw/202301/W020230118415217021121.pdf},
  note   = {Original title in Chinese; accessed Jul. 27, 2026},
}

@article{PegadoBardayo2023,
  author  = {Pegado-Bardayo, Ana and Lorenzo-Espejo, Antonio and Mu{\~n}uzuri, Jes{\'u}s and Aparicio-Ruiz, Pablo},
  title   = {A Data-Driven Decision Support System for Service Completion Prediction in Last Mile Logistics},
  journal = {Transportation Research Part A: Policy and Practice},
  volume  = {176},
  pages   = {103817},
  year    = {2023},
  doi={10.1016/j.tra.2023.103817},
}

@TechReport{JDAnnualReport2021,
  author      = {{JD.com, Inc.}},
  title       = {Annual Report 2021},
  institution = {JD.com, Inc.},
  year        = {2022},
  url={https://www1.hkexnews.hk/listedco/listconews/sehk/2022/0428/2022042803935.pdf},
  note        = {Accessed Jul. 27, 2026},
}

@Online{MeituanUAV2025,
  author = {{Meituan}},
  title  = {Meituan Strengthens Independent Innovation in {UAVs} and Artificial Intelligence},
  year   = {2025},
  month  = feb,
  url={https://www.meituan.com/news/NN250221067008264},
  note   = {Original title in Chinese; accessed Jul. 27, 2026},
}

@Online{FAAPart135DroneDelivery,
  author = {{Federal Aviation Administration}},
  title  = {Package Delivery by Drone ({Part 135})},
  year   = {2026},
  url={https://www.faa.gov/uas/advanced_operations/package_delivery_drone},
  note   = {Accessed Jul. 27, 2026},
}

@article{Wang2018UrbanTransfer,
  author  = {Wang, Leye and Guo, Bin and Yang, Qiang},
  title   = {Smart City Development with Urban Transfer Learning},
  journal = {Computer},
  volume  = {51},
  number  = {12},
  pages   = {32--41},
  year    = {2018},
  doi={10.1109/MC.2018.2880015},
}

@article{ALOTAIBI2025104213,
title = {Optimizing disaster response with UAV-mounted RIS and HAP-enabled edge computing in 6G networks},
journal = {Journal of Network and Computer Applications},
volume = {241},
pages = {104213},
year = {2025},
issn = {1084-8045},
doi={https://doi.org/10.1016/j.jnca.2025.104213},
url={https://www.sciencedirect.com/science/article/pii/S1084804525001109},
author = {Jamal Alotaibi and Omar Sami Oubbati and Mohammed Atiquzzaman and Fares Alromithy and Mohammad Rashed Altimania},
}

@article{xie2023transfloor,
  title={TransFloor: Transparent floor localization for crowdsourcing instant delivery},
  author={Xie, Zhiqing and Luo, Haiyong and Zhang, Xiaotian and Xiong, Hao and Zhao, Fang and Li, Zhaohui and Ye, Qi and Rong, Bojie and Gao, Jiuchong},
  journal={Proceedings of the ACM on Interactive, Mobile, Wearable and Ubiquitous Technologies},
  volume={6},
  number={4},
  pages={1--30},
  year={2023},
  publisher={ACM New York, NY, USA}
}

@inproceedings{ding2021nationwide,
  title={Nationwide deployment and operation of a virtual arrival detection system in the wild},
  author={Ding, Yi and Yang, Yu and Jiang, Wenchao and Liu, Yunhuai and He, Tian and Zhang, Desheng},
  booktitle={Proceedings of the 2021 ACM SIGCOMM 2021 Conference},
  pages={705--717},
  year={2021}
}

@article{kingma2014adam,
  title={Adam: A method for stochastic optimization},
  author={Kingma, Diederik P and Ba, Jimmy},
  journal={arXiv preprint arXiv:1412.6980},
  year={2014}
}

@article{kobusingye2006emergency,
  title={Emergency medical services},
  author={Kobusingye, Olive C and Hyder, Adnan A and Bishai, David and Joshipura, Manjul and Hicks, Eduardo Romero and Mock, Charles},
  journal={Disease Control Priorities in Developing Countries. 2nd edition},
  year={2006},
  publisher={The International Bank for Reconstruction and Development/The World Bank}
}

@inproceedings{jiang2023faircod,
  title={FairCod: A fairness-aware concurrent dispatch system for large-scale instant delivery services},
  author={Jiang, Lin and Wang, Shuai and Guo, Baoshen and Wang, Hai and Zhang, Desheng and Wang, Guang},
  booktitle={Proceedings of the 29th ACM SIGKDD Conference on knowledge discovery and data mining},
  pages={4229--4238},
  year={2023}
}

@TechReport{UAVCostPerMinute,
  author = {Kong, Rong and Yang, Yuchen},
  title  = {An Overview of {Meituan}'s {AI} and Robotics Initiatives: Has the Singularity Arrived for Autonomous Delivery?},
  institution = {TF Securities},
  type   = {Industry Research Report},
  year   = {2025},
  month  = mar,
  url={https://pdf.dfcfw.com/pdf/H3_AP202503111644266760_1.pdf},
  note   = {Original title in Chinese; accessed Jul. 27, 2026}
}

@TechReport{ZTO,
  author = {{ZTO Research Institute}},
  title  = {Cost Analysis of ``Last-mile + Feeder'' Drone Operational Scenarios},
  institution = {ZTO Research Institute},
  type   = {Industry Research Report},
  year   = {2021},
  url={https://hkjjyjzx.zua.edu.cn/info/1066/1264.htm},
  note   = {Hosted by the Henan Provincial Aviation Economics Research Center; original title in Chinese; accessed Jul. 27, 2026}
}

@inproceedings{10.1145/3372224.3419198,
author = {Yang, Yu and Ding, Yi and Yuan, Dengpan and Wang, Guang and Xie, Xiaoyang and Liu, Yunhuai and He, Tian and Zhang, Desheng},
title = {TransLoc: Transparent Indoor Localization with Uncertain Human Participation for Instant Delivery},
year = {2020},
isbn = {9781450370851},
publisher = {Association for Computing Machinery},
address = {New York, NY, USA},
doi={10.1145/3372224.3419198},
booktitle = {Proceedings of the 26th Annual International Conference on Mobile Computing and Networking},
articleno = {41},
numpages = {14},
location = {London, United Kingdom},
series = {MobiCom '20}
}

@article{LI201431,
title = {The Share-a-Ride Problem: People and parcels sharing taxis},
journal = {European Journal of Operational Research},
volume = {238},
number = {1},
pages = {31-40},
year = {2014},
issn = {0377-2217},
doi={https://doi.org/10.1016/j.ejor.2014.03.003},
author = {Baoxiang Li and Dmitry Krushinsky and Hajo A. Reijers and Tom {Van Woensel}}}

@article{wen2022joint,
  title={Joint Trajectory and Pick-Up Design for UAV-Assisted Item Delivery Under No-Fly Zone Constraints},
  author={Wen, Wanli and Luo, Kun and Liu, Lanhui and Zhang, Yi and Jia, Yunjian},
  journal={IEEE Transactions on Vehicular Technology},
  volume={72},
  number={2},
  pages={2587--2592},
  year={2022},
  publisher={IEEE}
}

@article{fatehi2022crowdsourcing,
  title={Crowdsourcing last-mile deliveries},
  author={Fatehi, Soraya and Wagner, Michael R},
  journal={Manufacturing \& Service Operations Management},
  volume={24},
  number={2},
  pages={791--809},
  year={2022},
  publisher={INFORMS}
}

@article{choudhury2021efficient,
  title={Efficient large-scale multi-drone delivery using transit networks},
  author={Choudhury, Shushman and Solovey, Kiril and Kochenderfer, Mykel J and Pavone, Marco},
  journal={Journal of Artificial Intelligence Research},
  volume={70},
  pages={757--788},
  year={2021}
}

@ARTICLE{7513397,
  author={Dorling, Kevin and Heinrichs, Jordan and Messier, Geoffrey G. and Magierowski, Sebastian},
  journal={IEEE Transactions on Systems, Man, and Cybernetics: Systems}, 
  title={Vehicle Routing Problems for Drone Delivery}, 
  year={2017},
  volume={47},
  number={1},
  pages={70-85},
  doi={10.1109/TSMC.2016.2582745}}

@Online{UAVRegulation,
  author = {{General Office of the Shanghai Municipal People's Government}},
  title  = {Interim Measures of {Shanghai Municipality} for the Flight Safety Management of Civil Unmanned Aircraft},
  year   = {2025},
  month  = dec,
  url={https://www.shanghai.gov.cn/nw12344/20251231/9d1638d5d8d44e67a02a46e6bb5a5eb7.html},
  note   = {Original title in Chinese; published Dec. 31, 2025; accessed Jul. 27, 2026}
}

@article{sun2019online,
  title={Online delivery route recommendation in spatial crowdsourcing},
  author={Sun, Dezhi and Xu, Ke and Cheng, Hao and Zhang, Yuanyuan and Song, Tianshu and Liu, Rui and Xu, Yi},
  journal={World Wide Web},
  volume={22},
  pages={2083--2104},
  year={2019},
  publisher={Springer}
}

@article{basik2018fair,
  title={Fair task allocation in crowdsourced delivery},
  author={Bas{\i}k, Fuat and Gedik, Bu{\u{g}}ra and Ferhatosmano{\u{g}}lu, Hakan and Wu, Kun-Lung},
  journal={IEEE Transactions on Services Computing},
  volume={14},
  number={4},
  pages={1040--1053},
  year={2018},
  publisher={IEEE}
}

@article{pan2021efficient,
  title={Efficient schedule of energy-constrained UAV using crowdsourced buses in last-mile parcel delivery},
  author={Pan, Yan and Li, Shining and Chen, Qianwu and Zhang, Nan and Cheng, Tao and Li, Zhigang and Guo, Bin and Han, Qingye and Zhu, Ting},
  journal={Proceedings of the ACM on Interactive, Mobile, Wearable and Ubiquitous Technologies},
  volume={5},
  number={1},
  pages={1--23},
  year={2021},
  publisher={ACM New York, NY, USA}
}

@article{cai2020500,
  title={A 500-W wireless charging system with lightweight pick-up for unmanned aerial vehicles},
  author={Cai, Chunwei and Wu, Shuai and Jiang, Longyun and Zhang, Zhipeng and Yang, Shiyan},
  journal={IEEE Transactions on Power Electronics},
  year={2020},
  publisher={IEEE}
}

@article{chen2022image,
  title={Image-based Visual Servoing of Unmanned Aerial Manipulators for Tracking and Grasping a Moving Target},
  author={Chen, Yanjie and Wu, Yangning and Zhang, Zhenguo and Miao, Zhiqiang and Zhong, Hang and Zhang, Hui and Wang, Yaonan},
  journal={IEEE Transactions on Industrial Informatics},
  year={2022},
  publisher={IEEE}
}

@article{dissanayaka2023review,
  title={Review of Navigation Methods for UAV-Based Parcel Delivery},
  author={Dissanayaka, Didula and Wanasinghe, Thumeera R and De Silva, Oscar and Jayasiri, Awantha and Mann, George KI},
  journal={IEEE Transactions on Automation Science and Engineering},
  year={2023},
  publisher={IEEE}
}

@inproceedings{xu2022drive,
  title={Drive Less but Finish More: Food Delivery based on Multi-Level Workers in Spatial Crowdsourcing},
  author={Xu, Xiaojia and Liu, An and Liu, Guanfeng and Li, Zhixu and Zhao, Lei},
  booktitle={Proceedings of the 31st ACM International Conference on Information \& Knowledge Management},
  pages={2331--2340},
  year={2022}
}

@article{ding2021city,
  title={A city-wide crowdsourcing delivery system with reinforcement learning},
  author={Ding, Yi and Guo, Baoshen and Zheng, Lin and Lu, Mingming and Zhang, Desheng and Wang, Shuai and Son, Sang Hyuk and He, Tian},
  journal={Proceedings of the ACM on Interactive, Mobile, Wearable and Ubiquitous Technologies},
  volume={5},
  number={3},
  pages={1--22},
  year={2021},
  publisher={ACM New York, NY, USA}
}

@article{chen2020measuring,
  title={Measuring maximum urban capacity of taxi-based logistics},
  author={Chen, Yueyue and Guo, Deke and Xu, Ming and Tang, Guoming and Cheng, Geyao},
  journal={IEEE Transactions on Intelligent Transportation Systems},
  volume={22},
  number={10},
  pages={6449--6459},
  year={2020},
  publisher={IEEE}
}

@InProceedings{10.1007/978-3-030-18579-4_45,
author="Cheng, Bin
and Qian, Shiyou
and Cao, Jian
and Xue, Guangtao
and Yu, Jiadi
and Zhu, Yanmin
and Li, Minglu
and Zhang, Tao",
editor="Li, Guoliang
and Yang, Jun
and Gama, Joao
and Natwichai, Juggapong
and Tong, Yongxin",
title="{STL}: Online Detection of Taxi Trajectory Anomaly Based on Spatial-Temporal Laws",
booktitle="Database Systems for Advanced Applications",
year="2019",
publisher="Springer International Publishing",
address="Cham",
pages="764--779",
isbn="978-3-030-18579-4"
}

@article{liu2018foodnet,
  title={FooDNet: Toward an optimized food delivery network based on spatial crowdsourcing},
  author={Liu, Yan and Guo, Bin and Chen, Chao and Du, He and Yu, Zhiwen and Zhang, Daqing and Ma, Huadong},
  journal={IEEE Transactions on Mobile Computing},
  volume={18},
  number={6},
  pages={1288--1301},
  year={2018},
  publisher={IEEE}
}

@article{devari2017crowdsourcing,
  title={Crowdsourcing the last mile delivery of online orders by exploiting the social networks of retail store customers},
  author={Devari, Aashwinikumar and Nikolaev, Alexander G and He, Qing},
  journal={Transportation Research Part E: Logistics and Transportation Review},
  volume={105},
  pages={105--122},
  year={2017},
  publisher={Elsevier}
}

@article{wang2022recommending,
  title={Recommending-and-Grabbing: A Crowdsourcing-Based Order Allocation Pattern for On-Demand Food Delivery},
  author={Wang, Xing and Wang, Ling and Wang, Shengyao and Pan, Jize and Ren, Hao and Zheng, Jie},
  journal={IEEE Transactions on Intelligent Transportation Systems},
  volume={24},
  number={1},
  pages={838--853},
  year={2022},
  publisher={IEEE}
}

@article{huang2020method,
  title={A method of optimized deployment of charging stations for drone delivery},
  author={Huang, Hailong and Savkin, Andrey V},
  journal={IEEE Transactions on Transportation Electrification},
  volume={6},
  number={2},
  pages={510--518},
  year={2020},
  publisher={IEEE}
}

@INPROCEEDINGS{10184563,
  author={Zhu, Guanzhou and Zhao, Dong and Wang, Yizong and Wang, Haotian and Zhang, Desheng and Ma, Huadong},
  booktitle={2023 IEEE 39th International Conference on Data Engineering (ICDE)}, 
  title={COME: Learning to Coordinate Crowdsourcing and Regular Couriers for Offline Delivery During Online Mega Sale Days}, 
  year={2023},
  volume={},
  number={},
  pages={3126-3139},
  doi={10.1109/ICDE55515.2023.00240}}

@ARTICLE{9524841,
  author={Ding, Yi and Liu, Ling and Yang, Yu and Liu, Yunhuai and Zhang, Desheng and He, Tian},
  journal={IEEE/ACM Transactions on Networking}, 
  title={From Conception to Retirement: A Lifetime Story of a 3-Year-Old Wireless Beacon System in the Wild}, 
  year={2022},
  volume={30},
  number={1},
  pages={47-61},
  doi={10.1109/TNET.2021.3107043}
  }

@article{pan2024pioneering,
  title={Pioneering Cooperative Air-Ground Instant Delivery Using UAVs and Crowdsourced Couriers},
  author={Pan, Yan and Gao, Junhui and Duan, Jiashun and Shi, Juan and Guo, Bin and Liang, Yunji and Hu, Yujiao},
  journal={Proceedings of the ACM on Interactive, Mobile, Wearable and Ubiquitous Technologies},
  volume={8},
  number={4},
  pages={1--26},
  year={2024},
  publisher={ACM New York, NY, USA}
}

@ARTICLE{11224006,
  author={Ameur, Abdelkader Ilyes and Oubbati, Omar Sami and Rachedi, Abderrezak and Arishi, Ali and Atiquzzaman, Mohammed},
  journal={IEEE Transactions on Network Science and Engineering}, 
  title={Intelligent UAV Caching and Energy Management in 6G Networks}, 
  year={2026},
  volume={13},
  number={},
  pages={3175-3192},
  doi={10.1109/TNSE.2025.3628171}}

@ARTICLE{9248613,
  author={Han, Yitao and Liu, Liang and Duan, Lingjie and Zhang, Rui},
  journal={IEEE Transactions on Wireless Communications}, 
  title={Towards Reliable UAV Swarm Communication in D2D-Enhanced Cellular Networks}, 
  year={2021},
  volume={20},
  number={3},
  pages={1567-1581},
  doi={10.1109/TWC.2020.3034457}}

@article{campion2018uav,
  title={UAV swarm communication and control architectures: a review},
  author={Campion, Mitch and Ranganathan, Prakash and Faruque, Saleh},
  journal={Journal of Unmanned Vehicle Systems},
  volume={7},
  number={2},
  pages={93--106},
  year={2018},
  publisher={NRC Research Press}
}

@ARTICLE{9756923,
  author={Pan, Yan and Chen, Qianwu and Zhang, Nan and Li, Zhigang and Zhu, Ting and Han, Qingye},
  journal={IEEE Transactions on Mobile Computing}, 
  title={Extending Delivery Range and Decelerating Battery Aging of Logistics UAVs Using Public Buses}, 
  year={2023},
  volume={22},
  number={9},
  pages={5280-5295},
  doi={10.1109/TMC.2022.3167040}}

@INPROCEEDINGS{10598004,
  author={Gao, Junhui and Wang, Qianru and Zhang, Xin and Shi, Juan and Zhao, Xiang and Han, Qingye and Pan, Yan},
  booktitle={2024 IEEE 40th International Conference on Data Engineering (ICDE)}, 
  title={Cooperative Air-Ground Instant Delivery by UAVs and Crowdsourced Taxis}, 
  year={2024},
  volume={},
  number={},
  pages={4153-4166},
  doi={10.1109/ICDE60146.2024.00120}}

@ARTICLE{7575702,
  author={Chen, Chao and Zhang, Daqing and Ma, Xiaojuan and Guo, Bin and Wang, Leye and Wang, Yasha and Sha, Edwin},
  journal={IEEE Transactions on Intelligent Transportation Systems}, 
  title={crowddeliver: Planning City-Wide Package Delivery Paths Leveraging the Crowd of Taxis}, 
  year={2017},
  volume={18},
  number={6},
  pages={1478-1496},
  doi={10.1109/TITS.2016.2607458}}

@ARTICLE{9151388,
  author={Huang, Hailong and Savkin, Andrey V. and Huang, Chao},
  journal={IEEE Transactions on Industrial Informatics}, 
  title={Drone Routing in a Time-Dependent Network: Toward Low-Cost and Large-Range Parcel Delivery}, 
  year={2021},
  volume={17},
  number={2},
  pages={1526-1534},
  doi={10.1109/TII.2020.3012162}}

@article{duan2023velp,
  title={Velp: vehicle loading plan learning from human behavior in nationwide logistics system},
  author={Duan, Sijing and Lyu, Feng and Zhu, Xin and Ding, Yi and Wang, Haotian and Zhang, Desheng and Liu, Xue and Zhang, Yaoxue and Ren, Ju},
  journal={Proceedings of the VLDB Endowment},
  volume={17},
  number={2},
  pages={241--249},
  year={2023},
  publisher={VLDB Endowment}
}

@ARTICLE{10976414,
  author={Oubbati, Omar Sami and Alotaibi, Jamal and Alromithy, Fares and Atiquzzaman, Mohammed and Altimania, Mohammad Rashed},
  journal={IEEE Transactions on Vehicular Technology}, 
  title={A UAV-UGV Cooperative System: Patrolling and Energy Management for Urban Monitoring}, 
  year={2025},
  volume={74},
  number={9},
  pages={13521-13536},
  doi={10.1109/TVT.2025.3563971}}

@ARTICLE{7876982,
  author={Ford, Russell and Zhang, Menglei and Mezzavilla, Marco and Dutta, Sourjya and Rangan, Sundeep and Zorzi, Michele},
  journal={IEEE Communications Magazine}, 
  title={Achieving Ultra-Low Latency in 5G Millimeter Wave Cellular Networks}, 
  year={2017},
  volume={55},
  number={3},
  pages={196-203},
  doi={10.1109/MCOM.2017.1600407CM}}

@ARTICLE{10167760,
  author={Bai, Xiaoshan and Ye, Youqiang and Zhang, Bo and Ge, Shuzhi Sam},
  journal={IEEE Transactions on Intelligent Transportation Systems}, 
  title={Efficient Package Delivery Task Assignment for Truck and High Capacity Drone}, 
  year={2023},
  doi={10.1109/TITS.2023.3287163}}

@Online{DJIMavic3Pro,
  author = {DJI},
  title  = {Characteristics of DJI Mavic 3 Pro},
  url={https://www.dji.com/cn/mavic-3-pro/specs}
}

@article{du2021uav,
  title={Uav trajectory planning with probabilistic geo-fence via iterative chance-constrained optimization},
  author={Du, Bin and Chen, Jun and Sun, Dengfeng and Manyam, Satyanarayana Gupta and Casbeer, David W},
  journal={IEEE Transactions on Intelligent Transportation Systems},
  volume={23},
  number={6},
  pages={5859--5870},
  year={2021},
  publisher={IEEE}
}

@article{auad2024dynamic,
  title={Dynamic courier capacity acquisition in rapid delivery systems: A deep q-learning approach},
  author={Auad, Ramon and Erera, Alan and Savelsbergh, Martin},
  journal={Transportation Science},
  volume={58},
  number={1},
  pages={67--93},
  year={2024},
  publisher={INFORMS}
}

@article{BOZANTA2022107871,
title = {Courier routing and assignment for food delivery service using reinforcement learning},
journal = {Computers And Industrial Engineering},
volume = {164},
pages = {107871},
year = {2022},
issn = {0360-8352},
doi={https://doi.org/10.1016/j.cie.2021.107871},
author = {Aysun Bozanta and Mucahit Cevik and Can Kavaklioglu and Eray M. Kavuk and Ayse Tosun and Sibel B. Sonuc and Alper Duranel and Ayse Basar},
}

@article{chen2024courier,
  title={Courier dispatch in on-demand delivery},
  author={Chen, Mingliu and Hu, Ming},
  journal={Management Science},
  volume={70},
  number={6},
  pages={3789--3807},
  year={2024},
  publisher={INFORMS}
}

@inproceedings{10.1145/3580305.3599766,
author = {Xia, Kaiwen and Lin, Li and Wang, Shuai and Wang, Haotian and Zhang, Desheng and He, Tian},
title = {A Predict-Then-Optimize Couriers Allocation Framework for Emergency Last-mile Logistics},
year = {2023},
isbn = {9798400701030},
publisher = {Association for Computing Machinery},
address = {New York, NY, USA},
doi={10.1145/3580305.3599766},
booktitle = {Proceedings of the 29th ACM SIGKDD Conference on Knowledge Discovery and Data Mining},
pages = {5237--5248},
numpages = {12},
location = {Long Beach, CA, USA},
series = {KDD '23}
}

@ARTICLE{8974403,
  author={Xu, Dongfang and Sun, Yan and Ng, Derrick Wing Kwan and Schober, Robert},
  journal={IEEE Transactions on Communications}, 
  title={Multiuser MISO UAV Communications in Uncertain Environments With No-Fly Zones: Robust Trajectory and Resource Allocation Design}, 
  year={2020},
  volume={68},
  number={5},
  pages={3153-3172},
  doi={10.1109/TCOMM.2020.2970043}}

@article{behrend2019exact,
  title={An exact solution method for the capacitated item-sharing and crowdshipping problem},
  author={Behrend, Moritz and Meisel, Frank and Fagerholt, Kjetil and Andersson, Henrik},
  journal={European Journal of Operational Research},
  volume={279},
  number={2},
  pages={589--604},
  year={2019},
  publisher={Elsevier}
}

@INPROCEEDINGS{7989675,
  author={Gawel, Abel and Kamel, Mina and Novkovic, Tonci and Widauer, Jakob and Schindler, Dominik and von Altishofen, Benjamin Pfyffer and Siegwart, Roland and Nieto, Juan},
  booktitle={2017 IEEE International Conference on Robotics and Automation (ICRA)}, 
  title={Aerial picking and delivery of magnetic objects with MAVs}, 
  year={2017},
  volume={},
  number={},
  pages={5746-5752},
  doi={10.1109/ICRA.2017.7989675}}

@Online{NFZDJI,
  author = {{DJI Fly Safe}},
  title = {No-Fly Zone Query},
  url={https://fly-safe.dji.com/nfz/nfz-query},
  note   = {Accessed: Jul. 27, 2026}
}

@ARTICLE{11261388,
  author={Gao, Junhui and Wang, Qianru and Zhang, Xin and Shi, Juan and Zhao, Xiang and Liang, Yunji and Guo, Bin and Han, Qingye and Pan, Yan},
  journal={IEEE Transactions on Mobile Computing}, 
  title={Cooperative Air-Ground Instant Delivery by UAVs and Crowdsourced Taxis: Joint UAV Station Deployment and Delivery Scheduling}, 
  year={2025},
  volume={},
  number={},
  pages={1-17},
  doi={10.1109/TMC.2025.3634430}}

@article{Deng_2026,
   title={{UAV}-enabled Computing Power Networks: Design and Performance Analysis under Energy Constraints},
   ISSN={2161-9875},
   url={http://dx.doi.org/10.1109/TMC.2026.3655118},
   DOI={10.1109/tmc.2026.3655118},
   journal={IEEE Transactions on Mobile Computing},
   publisher={Institute of Electrical and Electronics Engineers (IEEE)},
   author={Deng, Yiqin and Fang, Zhengru and Hu, Senkang and Ma, Yanan and Guo, Xiaoyu and Zhang, Haixia and Fang, Yuguang},
   year={2026},
   pages={1--17} }

@article{COHEN2006162,
title = {An efficient approximation for the Generalized Assignment Problem},
journal = {Information Processing Letters},
volume = {100},
number = {4},
pages = {162-166},
year = {2006},
issn = {0020-0190},
doi={https://doi.org/10.1016/j.ipl.2006.06.003},
author = {Reuven Cohen and Liran Katzir and Danny Raz},
}

@article{xiang2021reusing,
  title={Reusing delivery drones for urban crowdsensing},
  author={Xiang, Chaocan and Zhou, Yanlin and Dai, Haipeng and Qu, Yuben and He, Suining and Chen, Chao and Yang, Panlong},
  journal={IEEE Transactions on Mobile Computing},
  year={2021},
  publisher={IEEE}
}

@article{sungur2010model,
  title={A model and algorithm for the courier delivery problem with uncertainty},
  author={Sungur, Ilgaz and Ren, Yingtao and Ord{\'o}{\~n}ez, Fernando and Dessouky, Maged and Zhong, Hongsheng},
  journal={Transportation science},
  volume={44},
  number={2},
  pages={193--205},
  year={2010},
  publisher={INFORMS}
}

@INPROCEEDINGS{10184811,
  author={Lyu, Wenjun and Wang, Haotian and Hong, Zhiqing and Wang, Guang and Yang, Yu and Liu, Yunhuai and Zhang, Desheng},
  booktitle={2023 IEEE 39th International Conference on Data Engineering (ICDE)}, 
  title={REDE: Exploring Relay Transportation for Efficient Last-mile Delivery}, 
  year={2023},
  volume={},
  number={},
  pages={3003-3016},
  doi={10.1109/ICDE55515.2023.00230}}

@Online{ShanghaiTaxiFee,
  author = {{Shanghai Municipal Transportation Commission}},
  title  = {Taxi Fare Structure and Charging Standards in {Shanghai}},
  year   = {2023},
  month  = may,
  url={https://jtw.sh.gov.cn/czqcyj/20180605/0010-10460.html},
  note   = {Original title in Chinese; accessed Jul. 27, 2026}
}

@TechReport{MeituanFeeStructure,
  author      = {Ting Wang},
  title       = {{Meituan-W}: A Leader in Local Services Expanding Boundaries through Retail and Technology},
  institution = {Essence International Securities (Hong Kong) Limited},
  type        = {Equity Research Report},
  year        = {2023},
  month       = oct,
  url={https://pdf.dfcfw.com/pdf/H3_AP202310161601476269_1.pdf},
  note        = {Original title in Chinese; accessed Jul. 27, 2026}
}

@Online{MeituanFulfillmentFee,
  author = {{Meituan}},
  title  = {What Is the Actual Commission Rate of {Meituan Waimai}?},
  year   = {2024},
  month  = dec,
  url={https://www.meituan.com/news/NN241225051008799},
  note   = {Original title in Chinese; accessed Jul. 27, 2026}
}

\end{document}